\documentclass[11pt]{article}

\IfFileExists{neurips_2026.sty}{\usepackage[eandd]{neurips_2026}}{}

\usepackage[utf8]{inputenc}
\usepackage[T1]{fontenc}
\usepackage[margin=1in]{geometry}
\usepackage{hyperref}
\usepackage{url}
\usepackage{booktabs}
\usepackage{amsfonts}
\usepackage{amsmath}
\usepackage{amssymb}
\usepackage{microtype}
\usepackage{graphicx}
\usepackage{xcolor}
\usepackage{multirow}
\usepackage{array}
\usepackage{longtable}
\usepackage{enumitem}
\usepackage[numbers]{natbib}
\newcolumntype{L}[1]{>{\raggedright\arraybackslash}p{#1}}
\newcommand{\BudgetBench}{\textsc{BudgetBench}}
\newcommand{\RunModel}{\texttt{qwen2.5:1.5b}}
\newcommand{\RunId}{\texttt{final\_qwen25\_15b\_89items}}
\newcommand{\OllamaEndpoint}{\url{http://localhost:11434/v1/chat/completions}}
\newcommand{\RepoURL}{\url{https://github.com/aviskaar/budgetbench}}

\title{\BudgetBench: A Budget-Tiered Protocol and Pilot Harness for Memory Strategy Evaluation in Local Large Language Model Agents}

\author{%
  Aditya Karnam Gururaj Rao\\
  \texttt{adityakarnam.grao@gmail.com}\\[0.6em]
  Arjun Jaggi\\
  \texttt{arjunjaggi@gmail.com}\\
}
\date{}

\begin{document}

\maketitle

\begin{abstract}
For local large language model agents, active context is a scarce operational
resource: memory capacity, prefill latency, cache growth, and service
objectives all constrain how many input tokens each call can afford.  We
present \BudgetBench{}, an active-budget evaluation protocol and reference
harness that treats the per-call input-token budget as the independent variable
when comparing memory strategies.  Holding the model, task, sampler, and
decoding parameters fixed, it sweeps budgets over 2K, 4K, 8K, 16K, and 32K
tokens and records quality, budget utilization, latency, and, as a first-class
outcome, budget-violation rates.  The core contribution is this reusable
measurement surface: a swappable \texttt{MemoryStrategy} contract, explicit
budget enforcement, deterministic or versioned graders, prompt-audit metadata,
and reproducibility artifacts, all released at \RepoURL{}.

We substantiate the protocol with pilot studies rather than final strategy
rankings.  Across a local \RunModel{} pilot (89 items each on SWE-bench
Verified and LongBench v2), a hosted 50-item Qwen3 30B-A3B LongBench
replication with exact tokenization, and a 500-item LongMemEval oracle study
scored by the official GPT-4o evaluator, the harness reliably exposes
budget-compliance failures, non-monotonic quality curves, and operating points
that single-budget evaluation hides.  The budgeted-versus-full-context
direction, however, remains unresolved: the local slice is near-null, the
hosted replication favors full context in point estimate, and no paired
interval establishes equivalence.  We report these results transparently,
including the finding that the early local pilot's tokenizer approximation
undercounts some served-model prompts, so its violation rows are
tokenizer-approximation diagnostics rather than claim-bearing compliance
results; all timings are operational diagnostics.  The reusable contribution is
the protocol, harness, and failure-reporting discipline needed to scale
fixed-budget memory-strategy evaluation into a community benchmark.
\end{abstract}

\section{Introduction}

Recent language models increasingly advertise context windows containing tens
or hundreds of thousands of tokens.  This trend can make explicit memory
management appear unnecessary: if the model can accept the entire trace, the
choice among retrieval, summarization, and truncation may seem secondary.
Local deployment conditions contradict this assumption.  A workstation,
laptop, or edge device may run an 8B or 14B model comfortably at short context
lengths but become slow or memory-constrained at longer lengths.  Even when a
long context fits in memory, an application may not tolerate the associated
prefill latency, memory pressure, or energy cost.  For local agents, active
context is an operational resource.

\BudgetBench{} is motivated by a practical question: \emph{given a fixed local
model, a fixed task, and a fixed active context budget, which memory strategy
should an agent use?}  The answer is not generally reducible to maximum context
length.  A truncation policy may be inexpensive but discard early evidence.  A
summary policy may preserve high-level state but introduce additional model
calls and summary drift.  Retrieval may preserve relevant evidence at tight
budgets but fail when the query is underspecified or when the retrieval corpus
is not chunked at a useful granularity.  Hierarchical memory systems such as
MemGPT-style stores and production systems such as Mem0 introduce additional
state and extraction policies, which may improve recall but also complicate
reproducibility~\citep{packer2023memgpt,chhikara2025mem0}.

Existing benchmarks address related problems.  SWE-bench evaluates whether
models can produce patches for real GitHub issues~\citep{jimenez2023swebench}.
LongBench and LongBench v2 evaluate long-context understanding across diverse
language tasks~\citep{bai2023longbench,bai2024longbenchv2}.  RULER and HELMET
stress-test long-context behavior across synthetic and application-centric
length scales~\citep{hsieh2024ruler,yen2024helmet}.  Lost-in-the-Middle shows
that accepting longer inputs does not imply reliable use of relevant
information throughout the sequence~\citep{liu2023lost}.  Agent benchmarks such
as $\tau$-bench evaluate tool-mediated interaction in realistic
domains~\citep{yao2024taubench}.  These benchmarks are valuable, but they do not
standardize the independent variable that local-agent developers routinely
control: the active context budget assigned to each model call.

Recent method papers make context budgets central, so \BudgetBench{} should
not be read as claiming that budget sweeping itself is new.  ContextBudget
formulates budget-aware context management as a sequential decision problem
for long-horizon search agents~\citep{wu2026contextbudget}.  BudgetMem learns
query-aware routing among budget-tiered memory
modules~\citep{zhang2026budgetmem}.  Engram reports that a lean retrieved context can
beat full history on LongMemEval\_S while using far fewer
tokens~\citep{wang2026engram}.  These papers raise the evidentiary bar: a benchmark
paper must compare against full context, report operational metrics, and make
its harness reusable.  The novelty claimed here is the conjunction of local
large language model (LLM) serving, fixed per-call active budgets, swappable
strategy implementations, versioned task graders, and comparable
quality-latency-compliance logs.

This paper contributes a budget-tiered protocol and a pilot implementation.
The implementation is intentionally modest.  It includes a
\texttt{MemoryStrategy} interface, a budget-enforced evaluation harness, JSON
Lines (JSONL) metric logging, and reference implementations for truncation,
summary-buffer compression, retrieval-augmented context selection, and an exposed
\texttt{full\_context} pass-through baseline for follow-up comparisons.  The
pilot run uses a small local model and 89 items per task to obtain a denser
pilot curve while remaining reproducible on a single machine.  The resulting
numbers should not be read as final rankings.  They provide an empirical
validation of the measurement surface: \BudgetBench{} can expose quality
curves, compliance failures, and runtime tradeoffs that would be invisible in
a single-budget evaluation.

\paragraph{Contributions.}
This work makes five contributions.  First, it defines active-context-budget
evaluation as a local-agent protocol distinct from standard long-context model
evaluation and from method-specific budget studies.  Second, it documents an
implementation whose strategy interface is designed to admit new memory policies without
modifying task logic.  Third, it reports an 89-item-per-task local pilot with complete
five-tier tables for quality, violation rate, budget usage, and duration.
Fourth, it adds a filtered small-model transfer check showing how the same
protocol is compatible with additional local text and vision-capable model
tags while keeping uninformative runs visible.  Fifth, it reports a complete
500-question LongMemEval oracle-file matrix with 5,122 upstream GPT-4o judge
labels and auditable per-cell artifacts.

\section{Protocol Definition}

\subsection{Objects and Notation}

Let $M$ denote a fixed language model exposed through a chat-completion
application programming interface (API).  Let $T$ denote a task distribution
with items $x \sim T$ and a versioned grader
$g(\hat{y}, x) \in [0,1]$, which may be deterministic or model-based.  Let $B$ be an active context budget measured
in tokens.  Let $\mathcal{S}$ be a set of memory strategies.  A strategy
$s \in \mathcal{S}$ maps an input message history $H = (m_1,\ldots,m_n)$ and
budget $B$ to a budget-constrained message list:
\[
  s(H, B) = H' \quad \text{such that} \quad \tau(H') \leq B,
\]
where $\tau$ is the benchmark tokenizer or tokenizer approximation.  The model
prediction is $\hat{y}=M(s(H, B))$, and the observed quality is $g(\hat{y}, x)$.

For $N$ evaluated items, the benchmark reports the empirical quality curve
\[
  Q(s, B, T, M) = \frac{1}{N}\sum_{i=1}^{N} g(M(s(H_i, B)), x_i),
\]
alongside utilization and reliability metrics.  The budget sweep is not a
model capability test in isolation; it is a controlled ablation over strategy
choice at fixed model, task, sampler, and decoding parameters.

\subsection{Active Budget Versus Nominal Context Window}

The active budget is the number of input tokens that the strategy is allowed
to present to the model for a particular call.  It is not the model's maximum
advertised context window, and it is not total task token cost.  This
distinction matters.  A model may support a 32K window while an application
allocates only 4K tokens to an interactive agent turn.  Conversely, an agent
may spend many total tokens across a task while each individual call remains
within a smaller active budget.  \BudgetBench{} isolates this per-call
constraint because that is the quantity memory strategies directly control.

\subsection{Rationale for Fixed Budgets}

Budget-tiered evaluation is useful for three reasons.  First, it creates
comparable operating points.  Reporting that a strategy performs well at
``short context'' is not precise; reporting quality at 2,048 tokens is.  Second,
it surfaces strategy crossovers.  Retrieval may dominate at 2K while
truncation may be sufficient at 16K.  Third, it reveals non-monotonic behavior.
Longer prompts can add distractors, shift relevant information away from
positions the model uses reliably, or trigger different generation behavior.
The long-context literature shows that larger context windows do not guarantee
reliable evidence use~\citep{liu2023lost,hsieh2024ruler}; \BudgetBench{}
extends this observation to memory strategy comparisons.

\section{Related Work}

\subsection{Closest Concurrent Work}

ContextBudget/BACM-RL is the closest budget-control method paper.  It
formalizes context management under a budget and evaluates learned compression
policies for long-horizon search agents~\citep{wu2026contextbudget}.  BudgetMem
also treats cost as a controllable axis, but routes among memory-module budget
tiers rather than exposing a raw per-call token budget and third-party strategy
contract~\citep{zhang2026budgetmem}.  MemoryArena shifts the focus toward
multi-session memory-agent-environment loops where memorization and action are
tightly coupled~\citep{he2026memoryarena}.  That benchmark is closer to the
long-horizon agent setting we eventually want, but it does not expose a general
fixed-token-tier protocol for third-party strategy plug-ins.  LongMemEval-V2
pushes memory evaluation toward environment experience and very long
multimodal trajectories, again without the present per-call budget sweep
framing~\citep{wu2026longmemevalv2}.  EvoMemBench benchmarks agent memory from
self-evolving in-episode and cross-episode perspectives, comparing many memory
methods with long-context baselines, but it is not primarily a fixed
active-budget harness for local deployments~\citep{wang2026evomembench}.
Engram is especially important because it includes full-context baselines,
reproducible logs, and evidence that lean retrieved context can achieve higher
accuracy than full history on LongMemEval\_S~\citep{wang2026engram}.  LightMem and Prompt
Compression in the Wild reinforce the operational lens: latency, online
overhead, hardware, and break-even points must be measured, not
assumed~\citep{zhang2026lightmem,kummer2026promptcompression}.

Table~\ref{tab:closest_work} summarizes how \BudgetBench{} differs from the
closest method and benchmark artifacts along the budget, strategy, deployment,
and metric axes.

\begin{table}[t]
\centering
\scriptsize
\setlength{\tabcolsep}{2pt}
\renewcommand{\arraystretch}{1.08}
\begin{tabular}{L{0.20\textwidth}L{0.15\textwidth}L{0.12\textwidth}L{0.13\textwidth}L{0.12\textwidth}L{0.15\textwidth}}
\toprule
Artifact & Primary focus & Fixed active budgets & Swappable strategies & Local-agent focus & Operational metrics \\
\midrule
\BudgetBench{} & Benchmark protocol & Yes & Yes & Yes & Quality, used/peak budget, violations, latency \\
ContextBudget~\citep{wu2026contextbudget} & Learned context-management method & Yes, internally & No & No & Quality and budget-dependent cost \\
BudgetMem~\citep{zhang2026budgetmem} & Query-aware memory-tier router & Tiered modules & No & Limited & Accuracy-cost frontiers \\
EvoMemBench~\citep{wang2026evomembench} & Agent-memory benchmark & No fixed tier protocol & Method comparisons & No & Memory-task quality by category \\
MemoryAgent\allowbreak Bench~\citep{hu2025memoryagentbench} & Incremental memory benchmark & No fixed tier protocol & Method comparisons & No & Memory competency scores \\
MemoryArena~\citep{he2026memoryarena} & Multi-session memory-agent loop benchmark & No fixed tier protocol & Method comparisons & Partial & Session-level task success \\
LongMemEval-V2~\citep{wu2026longmemevalv2} & Agent-environment memory benchmark & No fixed tier protocol & Method comparisons & Partial & Accuracy and query latency \\
Engram~\citep{wang2026engram} & Bi-temporal memory engine & Lean vs full context & No & Partial & Accuracy, tokens, errors, reproducible logs \\
RULER~\citep{hsieh2024ruler} & Long-context model capacity & Length sweep & No & No & Task accuracy by sequence length \\
HELMET~\citep{yen2024helmet} & Long-context model evaluation & Length sweep & No & No & Application-category quality \\
LongBench v2~\citep{bai2024longbenchv2} & Long-context QA substrate & Natural long inputs & No & No & Exact-match multiple-choice quality \\
\bottomrule
\end{tabular}
\caption{Positioning against the closest concurrent and background artifacts.  The intended distinction is the conjunction of fixed active budgets, strategy plug-ins, local deployment focus, deterministic graders, and operational metrics.}
\label{tab:closest_work}
\end{table}

\subsection{Long-Context Benchmarks}

LongBench introduces a unified bilingual benchmark for long-context
understanding across single-document QA, multi-document QA, summarization,
few-shot learning, synthetic tasks, and code completion~\citep{bai2023longbench}.
LongBench v2 shifts toward more demanding multiple-choice settings with long
contexts and reasoning-oriented questions~\citep{bai2024longbenchv2}.  These
benchmarks are useful substrates for \BudgetBench{} because they expose
long-context evidence selection while retaining automatic grading.  However,
their default evaluation protocol asks how well a model handles a given long
input, not how well a memory strategy adapts the input to a fixed active
budget.

RULER evaluates long-context models with configurable synthetic tasks including
retrieval, multi-hop tracing, aggregation, and question
answering~\citep{hsieh2024ruler}.  Its central lesson is that isolated retrieval success
can overstate long-context competence.  HELMET argues similarly for
application-centric evaluation across diverse long-context categories and
finds that synthetic tasks are not reliable predictors of downstream
performance~\citep{yen2024helmet}.  \BudgetBench{} is complementary: it does
not replace long-context benchmarks, but instead imposes a fixed-budget axis
on task families where memory strategies are expected to matter.

\subsection{Agentic and Software Engineering Benchmarks}

SWE-bench grounds coding-agent evaluation in real repository issues and
patches~\citep{jimenez2023swebench}.  SWE-bench Verified further filters the
benchmark to a human-validated subset of 500 instances.  In a full
\BudgetBench{} benchmark study, official containerized SWE-bench grading should
be used.
The current pilot uses the public task data but replaces the official oracle
with deterministic patch-similarity scoring so that the budget matrix can be
run quickly on a single workstation.

$\tau$-bench evaluates tool-agent-user interaction in retail and airline
domains with domain APIs, policy documents, simulated users, and state-based
grading~\citep{yao2024taubench}.  It is a strong candidate for future
\BudgetBench{} expansion because tool-use chains stress exactly the kind of
state retention, policy recall, and evidence selection that memory strategies
are designed to support.  The current repository contains a $\tau$-bench
integration path, but the pilot run omits it because the local tau2-bench data
dependency was not installed.

\subsection{Memory, Retrieval, and Compression}

Retrieval-augmented generation (RAG) combines parametric model knowledge with
non-parametric retrieval, improving factuality and specificity on
knowledge-intensive tasks~\citep{lewis2020rag}.  In the \BudgetBench{} setting,
retrieval is not merely a way to add external knowledge; it is a strategy for
selecting which internal task evidence enters a constrained prompt.  The RAG
baseline in this repository embeds intermediate messages, retrieves messages
relevant to the current query, and reconstructs a prompt that fits the active
budget.

MemGPT frames language-model memory as an operating-system-like hierarchy with
explicit management of context and archival stores~\citep{packer2023memgpt}.
Mem0 presents a production-oriented memory layer that extracts, consolidates,
and retrieves salient conversational facts with substantial latency and token
savings relative to full-context approaches~\citep{chhikara2025mem0}.  These
systems are commonly evaluated on chat-memory substrates such as LongMemEval
and LoCoMo, which test long-term interactive memory over timestamped
conversation histories or very long conversational traces~\citep{wu2024longmemeval,maharana2024locomo}.  Together, these systems motivate richer
strategy families than the three baselines used in the
pilot.  LLMLingua-2 represents a complementary compression family: rather than
choosing which chunks to include, it compresses prompt content toward a target
budget~\citep{pan2024llmlingua2}.  \BudgetBench{} is designed to accommodate
these systems under the same interface once their dependencies and local
configuration are available.

\section{Evaluation Protocol}

\subsection{Budget Tiers}

The default budget tiers are 2K, 4K, 8K, 16K, and 32K tokens.  They are
log-spaced enough to reveal broad scaling trends while remaining recognizable
deployment regimes for local models.  A benchmark run should report every
strategy at every tier, even if some cells fail, because budget-compliance
failures are part of the measurement.  Missing cells make Pareto analysis
ambiguous; explicit failures preserve information about which strategies cannot
operate under a requested constraint.

\subsection{Strategy Contract}

The strategy contract is deliberately narrow.  A strategy receives a list of
chat messages and an integer budget.  It returns a list of chat messages.  The
runner then enforces the budget independently.  This separation prevents a
strategy from silently exceeding budget and avoids task-specific special cases.
Strategies may be stateful within a task item, but the runner resets them
between items in the current pilot.  Resetting makes the pilot easier to
interpret: each item is evaluated independently rather than benefiting from
cross-item memory.

\subsection{Task Contract}

A task provides three methods: dataset loading, task execution, and grading.
Execution is responsible for formatting a benchmark item as messages, invoking
the shared evaluation runner, extracting a prediction, and returning it to the
grader.  Deterministic graders are preferred when they faithfully implement
the task metric.  When a public benchmark specifies a model-based evaluator,
the harness exports every prediction and records the exact judge model,
evaluator version, and row-level labels.  All cells in a comparison must use
the same grader configuration because judge variation can exceed the effects
under study.

\subsection{Metric Contract}

Each cell logs raw per-item events and aggregate summary rows.  The primary
quality metric depends on the task: patch-similarity for the current SWE
scaffold and exact-match accuracy for LongBench v2.  The reliability metric is
budget violation rate.  The resource metrics are mean used budget, maximum peak
budget, duration, and tokens per task.  The analyzer supports bootstrap
confidence intervals when item-level quality rows are available.  A full
benchmark should report those intervals and repeated trials; the present pilot
tables remain point-estimate artifact validation.

\section{Implementation Details}

\subsection{Repository Structure}

The implementation is a compact Python package under \path{src/budgetbench}.
Core budget enforcement lives in \path{budgetbench/core}.  The evaluation runner
and metric logger live in \path{budgetbench/evaluation}.  Task adapters live in
\path{budgetbench/tasks}.  Strategy implementations live in
\path{budgetbench/strategies}.  The user-facing driver is
\path{scripts/run_pilot.py}; analysis and plotting utilities are in
\path{scripts/analyze_results.py} and \path{scripts/plot_tradeoffs.py}.

\subsection{Budget Enforcement Path}

For each item, the runner resets the strategy, applies it to the formatted
messages, counts tokens, and raises a budget exception if the processed prompt
exceeds the active tier.  The runner retries budget processing up to a fixed
limit.  If all retries fail, it logs a failed metric row with violation rate
1.0 for that item.  The current runner can be configured with an explicit
tokenizer identifier and now logs prompt-audit artifacts for each processed
prompt: tokenizer metadata, a prompt hash, the post-strategy prompt payload,
and the counted input tokens before enforcement.  This design makes budget failures visible in the result
tables instead of hiding them as missing predictions.

\subsection{Logging and Analysis}

The raw logs are JSONL files, one per task-strategy-budget combination.  The
summary file records one aggregate row per cell.  The analysis script reads the
summary rows and then reopens per-cell logs to compute violation rate, mean
used budget, and maximum observed peak budget.  The analyzer supports both
older boolean \texttt{violation} fields and the current runner's numeric
\texttt{violation\_rate} rows.  This compatibility is important because a
table that reports all-zero violations while raw retry failures exist would
understate one of the key outcomes of budget-tiered evaluation.  Newer logs
also include item-level quality rows, allowing the analyzer to emit bootstrap
quality intervals without changing the raw logging format.  The execution
driver additionally supports repeated cell runs and shuffled cell order so
latency and stability measurements can be gathered without conflating them
with a fixed warmup schedule.

\subsection{Multimodal Boundary}

The current benchmark task adapters are text-first: LongBench v2, the SWE
scaffold, and the $\tau$-bench integration path all produce chat messages.
This matters for interpreting vision-capable local model tags.  A
vision-language model can be served through the same local stack, but a
text-only LongBench prompt is not a multimodal benchmark.  In this revision,
the Qwen-VL text-only LongBench cells are treated as an endpoint
check, while a separate six-item generated image QA set verifies the native
image-plus-text path.  A multimodal \BudgetBench{} extension should add
image or document-page payloads, visual question answering, chart/table
grounding, and multimodal retrieval policies under the same active-budget
discipline.

\section{Experimental Setup}

\subsection{Hardware and Serving}

The pilot was run locally on June 16, 2026 through Ollama's OpenAI-compatible
endpoint at \OllamaEndpoint.  The model was \RunModel{}, selected because it
could complete the full matrix within a practical runtime for local
experimentation.  The system also had larger MLX-backed models installed, but
those models produced reasoning-first responses through the local endpoint in
preliminary checks, which made them poorly suited for the exact-match pilot.
The \RunModel{} tag was installed locally through Ollama for the 89-item-per-task run.

After the primary 89-item-per-task run, we installed additional local Ollama tags:
\texttt{gemma4:e2b}, \texttt{qwen3.5:0.8b}, \texttt{qwen3.5:2b}, and
\texttt{qwen3-vl:2b}.  All four tags responded through the OpenAI-compatible
chat path in preliminary checks.  We then ran a compact three-item LongBench v2
transfer check over the 2K, 4K, and 8K tiers using truncation and RAG.  The
paper reports only the informative \texttt{gemma4:e2b} rows.  The Qwen text
tags and the Qwen-VL tag are preserved in raw logs, in the consolidated
comma-separated values (CSV) file, and in the zero-row disclosure table because
their compact text-only exact-match results were uniformly zero or required
endpoint-specific generation settings to avoid empty content.  We separately generated six simple colored-shape
images saved as portable network graphics (PNG) files and queried
\texttt{qwen3-vl:2b} through Ollama's native image chat API; that
native multimodal probe is reported because it produced nonzero results.

Only \RunModel{} is treated as a claim-bearing, publicly reproducible model
tag in this paper.  The supplementary tags above and the later local aliases
\texttt{qwen3.6:35b-mlx}, \texttt{gemma4:12b-mlx},
\texttt{gemma4:31b-mlx}, and \texttt{gemma3:1b} are local installation names
without source registry, quantization hash, or conversion-command provenance.
Their rows are therefore infrastructure and transfer diagnostics rather than
benchmark claims a reader is expected to reproduce exactly.

\subsection{Sampling and Decoding}

The run uses 89 items per task.  For exact-match tasks, 89 examples produce
accuracy increments of approximately 0.0112, reducing the coarse quantization
of smaller pilot samples.  The decoding temperature is 0.0, the maximum
generation length is 512 tokens, and the random seed is 42.  The evaluated
matrix contains two tasks, three strategies, and five budget tiers, yielding
30 task-strategy-budget cells.

\subsection{Tasks and Graders}

The SWE task uses items from SWE-bench Verified and asks the model to output a
unified diff.  The pilot grader computes a patch-similarity proxy:
40\% file-level recall plus 60\% changed-line overlap on matched files.  This
metric is deterministic and cheap, but it is not equivalent to official
SWE-bench resolved rate.

The LongBench v2 task formats the context, question, and four choices as chat
messages.  The context is chunked into retrievable message parts before memory
strategy processing.  The grader extracts an answer letter and performs
exact-match scoring.  This task provides the more direct pilot measurement because
it is deterministic and natively multiple-choice.

\subsection{Baselines}

The main 89-item pilot evaluates three baselines.  Truncation keeps the system prompt and
the newest messages that fit.  Summary-buffer compression summarizes evicted
context with the same local model and prepends that summary to the retained
messages.  RAG embeds intermediate context messages with a sentence-transformer
model, stores them in an in-memory Chroma collection, retrieves messages
relevant to the query, and reconstructs a prompt within the active budget.
The current harness also exposes a \texttt{full\_context} baseline for
follow-up runs.  It returns the task's natural formatted prompt unchanged and
lets the shared budget enforcer record whether that prompt fits the selected
tier, making infeasible full-context rows explicit.

\section{Results}

\begin{figure}[t]
  \centering
  \includegraphics[width=\textwidth]{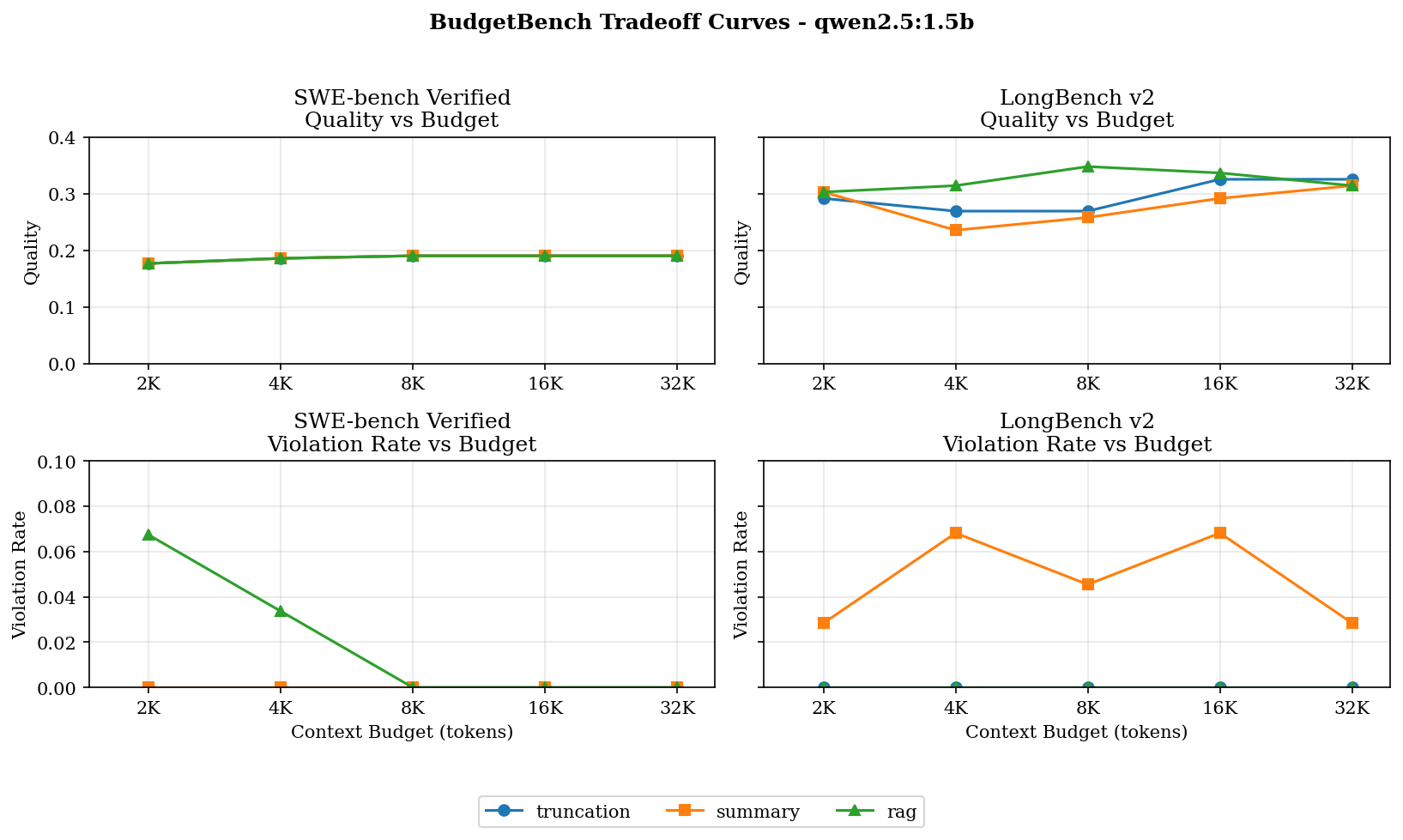}
  \caption{Quality and budget-violation curves for the 89-item-per-task pilot run. Blue circles denote truncation, orange squares denote summary, and green triangles denote retrieval-augmented generation (RAG). SWE quality is the patch-similarity proxy; LongBench v2 quality is exact-match accuracy.}
  \label{fig:tradeoff_89item}
\end{figure}

\begin{figure}[t]
  \centering
  \includegraphics[width=\textwidth]{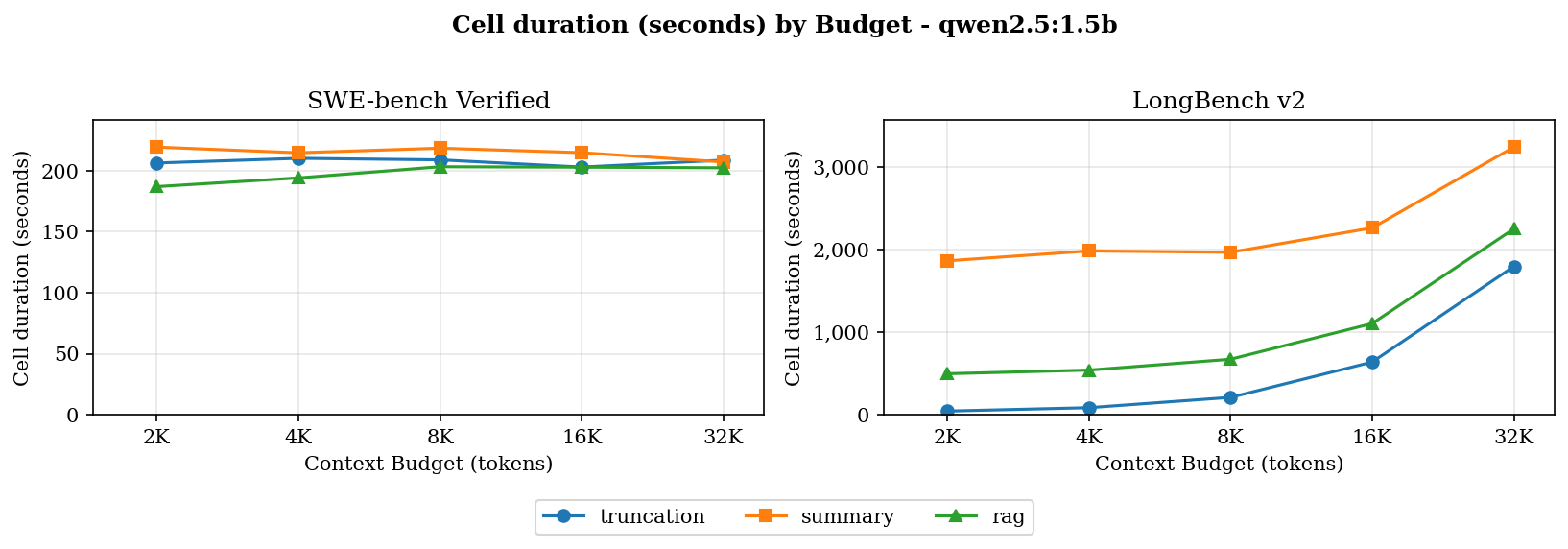}
  \caption{Cell-level runtime by task, strategy, and active context budget. Blue circles denote truncation, orange squares denote summary, and green triangles denote retrieval-augmented generation (RAG).}
  \label{fig:duration_89item}
\end{figure}

\begin{figure}[t]
  \centering
  \includegraphics[width=\textwidth]{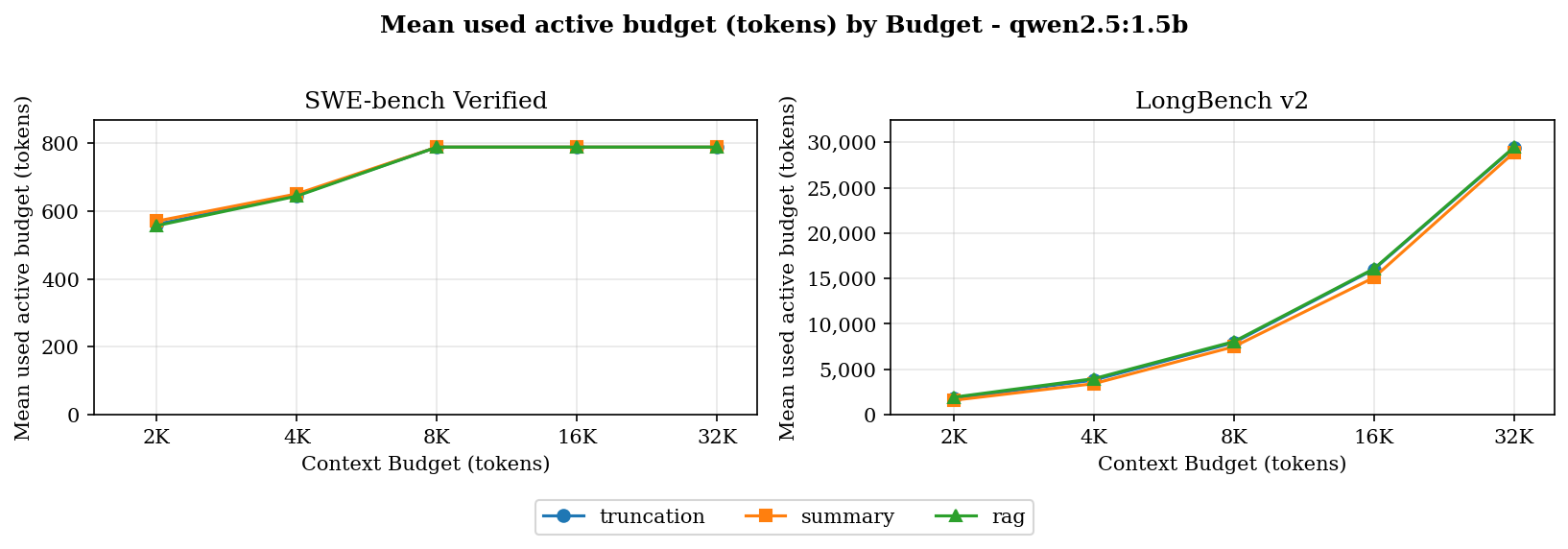}
  \caption{Mean used active budget by task, strategy, and tier. Blue circles denote truncation, orange squares denote summary, and green triangles denote retrieval-augmented generation (RAG).}
  \label{fig:used_budget_89item}
\end{figure}

This section first tests whether the reference strategies produce different
quality, violation, and runtime curves under fixed active-context budgets.  The
pilot artifacts are tracked in the public repository at \RepoURL{}.  The
aggregate CSV and figures are generated by the scripts in \path{scripts/}.
Table values are rounded to two decimals in the main text.  Table~\ref{tab:main_89item_results}
reports the 89-item LongBench v2 exploratory quality and tokenizer-approximation
violation matrix, while
Table~\ref{tab:swe_proxy_diagnostic} reports the SWE scaffold diagnostic.
Figures~\ref{fig:tradeoff_89item},~\ref{fig:duration_89item}, and~\ref{fig:used_budget_89item}
show the corresponding quality, violation,
duration, and budget-use curves.  The SWE cells are not official SWE-bench
accuracy, and the repeated high-budget proxy values occur because the selected
SWE prompts collapse into the same effective prompt regimes at the measured
budgets rather than because the strategies have independently identical coding
performance.  Therefore, the SWE scaffold is a plumbing check for budget
enforcement and logging, not evidence that memory strategy is irrelevant for
software engineering tasks.

\begin{table}[t]
\centering
\caption{LongBench v2 exploratory tokenizer-approximation diagnostics for \RunModel. Accuracy cells report exact-match accuracy with 95\% bootstrap confidence intervals computed from the available 88 item-level quality rows; point estimates are the aggregate run values. Violation rates are measured with the benchmark tokenizer approximation and should not be read as served-model-tokenizer compliance. Retrieval-augmented generation (RAG) denotes the episodic retrieval baseline.}
\label{tab:main_89item_results}
\resizebox{\textwidth}{!}{%
\begin{tabular}{llccccc}
\toprule
Strategy & Metric & 2K & 4K & 8K & 16K & 32K \\
\midrule
Truncation & Accuracy (95\% CI) & 0.29 [0.20, 0.40] & 0.27 [0.18, 0.36] & 0.27 [0.18, 0.36] & 0.33 [0.23, 0.43] & 0.33 [0.24, 0.43] \\
 & Violation rate & 0.00 & 0.00 & 0.00 & 0.00 & 0.00 \\
Summary & Accuracy (95\% CI) & 0.30 [0.22, 0.40] & 0.24 [0.16, 0.33] & 0.26 [0.17, 0.35] & 0.29 [0.20, 0.40] & 0.31 [0.23, 0.41] \\
 & Violation rate & 0.03 & 0.07 & 0.05 & 0.07 & 0.03 \\
RAG & Accuracy (95\% CI) & 0.30 [0.20, 0.40] & 0.31 [0.23, 0.42] & 0.35 [0.25, 0.45] & 0.34 [0.24, 0.43] & 0.31 [0.22, 0.42] \\
 & Violation rate & 0.00 & 0.00 & 0.00 & 0.00 & 0.00 \\
\bottomrule
\end{tabular}%
}
\end{table}

\begin{table}[t]
\centering
\caption{SWE-bench Verified scaffold diagnostic for the same 89-item-per-task run. Retrieval-augmented generation (RAG) denotes the episodic retrieval baseline. The SWE value is a patch-similarity proxy, not official SWE-bench accuracy. Repeated high-budget cells remain collapsed because these SWE prompts are short enough that the effective strategy inputs become identical at 8K and above.}
\label{tab:swe_proxy_diagnostic}
\resizebox{\textwidth}{!}{%
\begin{tabular}{llcccc}
\toprule
Strategy & Budget regime & SWE proxy & Violation rate & Mean used tokens & Max peak tokens \\
\midrule
Truncation & 2K & 0.18 & 0.00 & 561 & 1,973 \\
Truncation & 4K & 0.19 & 0.00 & 646 & 3,299 \\
Summary & 2K & 0.18 & 0.00 & 570 & 1,973 \\
Summary & 4K & 0.19 & 0.00 & 650 & 3,299 \\
RAG & 2K & 0.18 & 0.07 & 557 & 4,542 \\
RAG & 4K & 0.19 & 0.03 & 644 & 4,542 \\
All strategies & 8K--32K & 0.19 & 0.00 & 788 & 4,542 \\
\bottomrule
\end{tabular}%
}
\end{table}

\subsection{Full-Context-Feasible LongBench Slice}

The initial 89-item-per-task pilot does not answer whether a budgeted strategy can
match full context, because many natural LongBench v2 prompts exceed small
active tiers and some exceed 32K.  We therefore ran a follow-up slice on the
first 50 LongBench v2 training items whose natural formatted prompt contains
at most 32,768 benchmark tokens.  The run used \RunModel{}, the same local
Ollama endpoint, three strategies, and three active budgets: 2K, 8K, and 32K.
The artifacts are tracked in the public repository at \RepoURL{}.
Table~\ref{tab:full_context_feasible_slice} reports the feasible-slice
aggregate rows, and Table~\ref{tab:paired_full_context_deltas} reports paired
item-level deltas against the 32K full-context baseline.  After
appending an 8K \texttt{lean\_retrieval} row, the aggregate CSV is
\path{results/full_study_qwen2_5_1_5b_20260618_155102.csv}, and the paired
comparison table is
\path{results/paired_deltas_qwen2_5_1_5b_20260618_155102.csv}.
After the tokenizer-validity audit, we also reran a compact hardened slice on
10 LongBench v2 items with the explicit
\texttt{Qwen/Qwen2.5-1.5B-Instruct} tokenizer, shuffled cell order, and two
repeats per cell.  The repeat-aggregated CSV and paired-delta CSV are
\path{results/full_study_qwen2_5_1_5b_20260623_112431.csv} and
\path{results/paired_deltas_qwen2_5_1_5b_20260623_112431.csv}.  That hardened
slice is too small to replace the 50-item exploratory table below, but it does
remove the tokenizer-proxy ambiguity for the current harness path and adds
repeated, order-randomized runtime measurements.
We also ran two compact RAG-only ablations on the same 10-item slice: one swaps
the default \texttt{all-MiniLM-L6-v2} retrieval embeddings for
\texttt{sentence-transformers/all-mpnet-base-v2}, and one forces 256-token
LongBench context chunks instead of the default adaptive chunking rule.  The
comparison artifact is \path{results/ablation_rag_long10_20260623.csv}.  On
this slice, the forced 256-token chunks improve the 8K RAG mean accuracy from
0.50 to 0.60, while the heavier MPNet retrieval model does not improve
accuracy and materially increases mean duration at every tier.

\begin{table}[t]
\centering
\small
\begin{tabular}{llrrrr}
\toprule
Strategy & Budget & Accuracy & Violation & Used tokens & Duration (s) \\
\midrule
Truncation & 2K & 0.34 & 0.00 & 1,751 & 31.6 \\
Truncation & 8K & 0.26 & 0.00 & 7,961 & 128.3 \\
Truncation & 32K & 0.32 & 0.00 & 20,921 & 610.0 \\
RAG & 2K & 0.24 & 0.00 & 1,892 & 49.7 \\
RAG & 8K & 0.34 & 0.00 & 8,018 & 144.6 \\
RAG & 32K & 0.32 & 0.00 & 20,921 & 608.5 \\
Lean retrieval & 8K & 0.30 & 0.00 & 7,956 & 147.6 \\
Full context & 2K & 0.00 & 1.00 & 0 & 10.3 \\
Full context & 8K & 0.00 & 1.00 & 0 & 10.3 \\
Full context & 32K & 0.32 & 0.00 & 20,921 & 605.9 \\
\bottomrule
\end{tabular}
\caption{Fifty-item LongBench v2 slice filtered to items whose natural prompt fits within 32K tokens.  Retrieval-augmented generation (RAG) denotes the episodic retrieval baseline.  Full context at 2K and 8K is infeasible by construction; those rows fail budget enforcement before model calls.}
\label{tab:full_context_feasible_slice}
\end{table}

\begin{table}[t]
\centering
\small
\begin{tabular}{lrrr}
\toprule
Candidate versus full context at 32K & Paired $N$ & Mean delta & 95\% bootstrap CI \\
\midrule
Truncation at 2K & 50 & +0.02 & [-0.10, +0.14] \\
Truncation at 8K & 50 & -0.06 & [-0.14, +0.02] \\
RAG at 2K & 50 & -0.08 & [-0.18, 0.00] \\
RAG at 8K & 50 & +0.02 & [-0.08, +0.12] \\
RAG at 32K & 50 & 0.00 & [0.00, 0.00] \\
Lean retrieval at 8K & 50 & -0.02 & [-0.12, +0.08] \\
\bottomrule
\end{tabular}
\caption{Paired item-level exact-match deltas against the 32K full-context baseline on the feasible slice.  Retrieval-augmented generation (RAG) denotes the episodic retrieval baseline.  The 95\% bootstrap confidence interval (CI) is computed over paired item outcomes.  Positive values mean the budgeted strategy answered more items correctly than full context on the same item set.  This table is a difference estimate, not an equivalence test.}
\label{tab:paired_full_context_deltas}
\end{table}

This slice gives the first direct full-context comparison in the artifact, but
the selection rule should be kept in view: items were filtered to fit 32K full
context, so full-context infeasibility at smaller tiers is partly a property of
the slice construction rather than an independent discovery.
The result should be interpreted conservatively: no positive paired bootstrap
confidence interval excludes zero, and the interval is too wide to establish
equivalence.  The notable signal is operational rather than inferential.  At
$N=50$, RAG at 8K has no detectable paired accuracy difference from full
context at 32K.  The single-run timing values are lower for the shorter prompt
on this local setup, but they are not repeated warmup-controlled latency
estimates and should not be used as a strategy-speed claim.  Full context is also
categorically infeasible at 2K and 8K, whereas truncation and RAG produce
budget-compliant prompts at those tiers.  The initial lean retrieval baseline
succeeds operationally but does not improve over simple RAG on this slice.
This supports the core artifact motivation that active-budget evaluation
surfaces a
quality-compliance tradeoff that ordinary full-context evaluation
hides, including negative baseline results that should be reported rather than
silently discarded.  On the newer 10-item hardened slice, full context at 32K
averages 0.40 exact-match accuracy, truncation at 2K and 8K averages 0.50,
and RAG at 8K averages 0.50.  The paired deltas against full context at 32K
are +0.10 with bootstrap CIs of [0.00, +0.30] for truncation at 2K,
truncation at 8K, and RAG at 8K.  Those values should still be treated as
artifact-hardening evidence rather than benchmark ranking claims because the
slice is only 10 items wide.

\subsection{Hosted Stronger-Model LongBench Replication}

The local 35B feasibility attempts were too slow to complete the full
LongBench matrix on the available workstation.  We therefore separated model-
scale transfer from local-serving transfer and ran a hosted replication through
OpenRouter on June 24, 2026.  The candidate was
\path{qwen/qwen3-30b-a3b-instruct-2507}; token accounting used the explicit
\path{Qwen/Qwen3-30B-A3B-Instruct-2507} Hugging Face tokenizer rather than a
proxy.  The study reused the same 50-item, 32K-feasible selection rule and
evaluated four strategies at 8K and 32K.  Cell order was shuffled and each cell
was repeated twice at temperature zero with a 16-token output cap.  The 800
evaluations comprised 700 admitted model calls and 100 expected pre-inference
rejections from the two 8K full-context cells.  All admitted outputs were
parseable; the audit found no unexpected errors or approximate-tokenizer rows.

\begin{table}[t]
\centering
\small
\setlength{\tabcolsep}{3pt}
\begin{tabular}{llrrrr}
\toprule
Strategy & Budget & Accuracy & 95\% CI & Violation & Duration (s) \\
\midrule
Truncation & 8K & 0.33 & [0.21, 0.45] & 0.00 & $157.8 \pm 17.5$ \\
Truncation & 32K & 0.53 & [0.39, 0.67] & 0.00 & $209.7 \pm 0.3$ \\
RAG & 8K & 0.46 & [0.33, 0.60] & 0.00 & $163.4 \pm 14.9$ \\
RAG & 32K & 0.52 & [0.39, 0.65] & 0.00 & $220.9 \pm 21.1$ \\
Lean retrieval & 8K & 0.44 & [0.31, 0.58] & 0.00 & $181.5 \pm 3.5$ \\
Lean retrieval & 32K & 0.51 & [0.37, 0.64] & 0.00 & $225.9 \pm 0.3$ \\
Full context & 8K & 0.00 & [0.00, 0.00] & 1.00 & $47.3 \pm 0.5$ \\
Full context & 32K & 0.53 & [0.39, 0.67] & 0.00 & $225.5 \pm 1.1$ \\
\bottomrule
\end{tabular}
\caption{Repeat-aggregated stronger-model LongBench replication.  Accuracy is
the mean over two shuffled repeats; confidence intervals bootstrap the 50
item-level outcomes after repeat aggregation.  Duration is mean end-to-end
cell time plus or minus the sample standard deviation across two repeats.
Truncation at 32K reconstructs the same natural prompts as full context at 32K
yet differs by about 16 seconds, illustrating provider-variance dominance in
these wall-clock measurements.  The 8K full-context row contains budget-gate
time only and no hosted model calls.}
\label{tab:hosted_stronger_longbench}
\end{table}

\begin{table}[t]
\centering
\small
\begin{tabular}{lrrr}
\toprule
Candidate versus full context at 32K & Paired $N$ & Mean delta & 95\% bootstrap CI \\
\midrule
Truncation at 8K & 50 & -0.20 & [-0.34, -0.07] \\
RAG at 8K & 50 & -0.07 & [-0.17, +0.02] \\
Lean retrieval at 8K & 50 & -0.09 & [-0.20, +0.02] \\
Truncation at 32K & 50 & 0.00 & [0.00, 0.00] \\
RAG at 32K & 50 & -0.01 & [-0.04, +0.02] \\
Lean retrieval at 32K & 50 & -0.02 & [-0.05, 0.00] \\
\bottomrule
\end{tabular}
\caption{Paired item-level deltas for the hosted stronger-model replication,
after averaging each item's outcome across two repeats.  Intervals that include
zero do not establish equivalence; they indicate that this 50-item study does
not resolve a difference at the stated confidence level.}
\label{tab:hosted_stronger_longbench_deltas}
\end{table}

The stronger model preserves the central compliance tradeoff but sharpens the
quality result.  Truncation at 8K is 0.20 below 32K full context and its paired
interval excludes zero.  RAG at 8K narrows the point difference to 0.07 and
has a lower observed mean end-to-end cell time in this hosted diagnostic, but its
$[-0.17,+0.02]$ interval is too wide for either a no-loss or an equivalence
claim.  At 32K, truncation reconstructs the same natural prompts as full
context and matches it item-for-item after repeat aggregation; the two
retrieval variants are within two percentage points.  Hosted inference for the
full matrix cost \$0.9095.  Provider scheduling and network time are included
in duration, and the 32K truncation/full-context duration gap despite
item-identical prompts shows that wall-clock values are dominated by provider
variance.  The table is therefore a single-setup operational diagnostic, not a
local-hardware speed benchmark or a strategy-latency claim.

\subsection{Synthetic Memory-Agent Pilot}

We also added a deterministic memory-shaped task to test whether the harness
can evaluate agent memory behavior beyond long-context multiple choice.  Each
item contains a short event history with distractors and asks for an exact
answer about one of five categories: single-session facts, knowledge updates,
temporal reasoning, preferences, or abstention.  This task is synthetic and is
not a replacement for LongMemEval, LoCoMo, or other public memory-agent
benchmarks; its role here is to validate the protocol on memory-specific
phenomena with deterministic scoring.  The raw logs are in
\RepoURL{}; the aggregate, paired, and category CSV files are
\path{results/full_study_qwen2_5_1_5b_20260618_203210.csv},
\path{results/paired_deltas_qwen2_5_1_5b_20260618_203210.csv}, and
\path{results/grouped_memory_category_qwen2_5_1_5b_20260618_203210.csv}.
We also ran a \texttt{checkpoint\_context} baseline that keeps a compact
extractive checkpoint, a recent tail, and query-relevant evidence.  On the
same 30-item memory slice, it lands between truncation and lean retrieval:
0.53 at 512, 0.67 at 1,024, and 0.80 at 2,048.  It is comparable to
truncation at the tightest tier (5.3 s vs. 5.7 s at 512) and remains similar
to RAG and lean retrieval at 2,048 tokens.  It still gives a more realistic
practitioner workflow than either raw truncation or pure retrieval.
Table~\ref{tab:synthetic_memory_pilot} reports the aggregate rows,
Table~\ref{tab:synthetic_memory_deltas} reports paired deltas against full
context at 2,048 tokens, and Table~\ref{tab:synthetic_memory_categories}
reports the category split.

\begin{table}[t]
\centering
\small
\begin{tabular}{llrrrr}
\toprule
Strategy & Budget & Accuracy & Violation & Used tokens & Duration (s) \\
\midrule
Truncation & 512 & 0.43 & 0.00 & 500 & 5.7 \\
Truncation & 1,024 & 0.43 & 0.00 & 1,016 & 8.4 \\
Truncation & 2,048 & 0.80 & 0.00 & 1,812 & 13.4 \\
RAG & 512 & 0.73 & 0.00 & 500 & 18.8 \\
RAG & 1,024 & 0.80 & 0.00 & 1,014 & 22.1 \\
RAG & 2,048 & 0.80 & 0.00 & 1,812 & 3.2 \\
Lean retrieval & 512 & 0.80 & 0.00 & 500 & 15.6 \\
Lean retrieval & 1,024 & 0.83 & 0.00 & 1,018 & 19.1 \\
Lean retrieval & 2,048 & 0.80 & 0.00 & 1,812 & 3.1 \\
Checkpoint context & 512 & 0.53 & 0.00 & 479 & 5.3 \\
Checkpoint context & 1,024 & 0.67 & 0.00 & 968 & 7.7 \\
Checkpoint context & 2,048 & 0.80 & 0.00 & 1,812 & 3.2 \\
Full context & 512 & 0.00 & 1.00 & 0 & 0.1 \\
Full context & 1,024 & 0.00 & 1.00 & 0 & 0.1 \\
Full context & 2,048 & 0.80 & 0.00 & 1,812 & 3.1 \\
\bottomrule
\end{tabular}
\caption{Thirty-item synthetic memory-agent pilot.  Retrieval-augmented generation (RAG) denotes the episodic retrieval baseline.  Durations are single-run diagnostics; the non-monotonic timing values likely reflect warmup and execution-order effects and should not be read as stable latency estimates.  Full context fails at
512 and 1,024 tokens because the natural memory trace exceeds those active
budgets.}
\label{tab:synthetic_memory_pilot}
\end{table}

\begin{table}[t]
\centering
\small
\begin{tabular}{lrrr}
\toprule
Candidate versus full context at 2,048 & Paired $N$ & Mean delta & 95\% bootstrap CI \\
\midrule
Truncation at 512 & 30 & -0.37 & [-0.57, -0.17] \\
Truncation at 1,024 & 30 & -0.37 & [-0.57, -0.17] \\
RAG at 512 & 30 & -0.07 & [-0.17, 0.00] \\
RAG at 1,024 & 30 & 0.00 & [0.00, 0.00] \\
Lean retrieval at 512 & 30 & 0.00 & [0.00, 0.00] \\
Lean retrieval at 1,024 & 30 & +0.03 & [0.00, +0.10] \\
Checkpoint context at 512 & 30 & -0.27 & [-0.50, 0.00] \\
Checkpoint context at 1,024 & 30 & -0.13 & [-0.30, +0.03] \\
Checkpoint context at 2,048 & 30 & 0.00 & [0.00, 0.00] \\
Full context at 512 & 30 & -0.80 & [-0.93, -0.63] \\
Full context at 1,024 & 30 & -0.80 & [-0.93, -0.67] \\
\bottomrule
\end{tabular}
\caption{Paired item-level exact-match deltas against the 2,048-token
full-context baseline on the synthetic memory task.  Retrieval-augmented generation (RAG) denotes the episodic retrieval baseline.  The positive
\texttt{lean\_retrieval} row corresponds to one additional correct item and
should be treated as a pilot signal rather than a stable ranking claim.}
\label{tab:synthetic_memory_deltas}
\end{table}

\begin{table}[t]
\centering
\small
\setlength{\tabcolsep}{3pt}
\begin{tabular}{lrrrrr}
\toprule
Category & Trunc. 1,024 & RAG 1,024 & Lean 512 & Lean 1,024 & Full 2,048 \\
\midrule
Single fact & 0.00 & 1.00 & 1.00 & 1.00 & 1.00 \\
Temporal & 0.00 & 1.00 & 1.00 & 1.00 & 1.00 \\
Knowledge update & 1.00 & 1.00 & 1.00 & 1.00 & 1.00 \\
Abstention & 1.00 & 1.00 & 1.00 & 1.00 & 1.00 \\
Preference & 0.17 & 0.00 & 0.00 & 0.17 & 0.00 \\
\bottomrule
\end{tabular}
\caption{Category-level exact-match accuracy on the synthetic memory pilot,
six items per category.  Trunc. denotes truncation, retrieval-augmented generation (RAG) denotes episodic retrieval, Lean denotes lean retrieval, and Full denotes full context.}
\label{tab:synthetic_memory_categories}
\end{table}

The memory pilot gives a different failure profile from LongBench.  Tight
truncation loses early single-session facts and temporal evidence, while
retrieval-based strategies recover those categories at tight budgets.  The
category split also prevents a false positive: preference questions remain
poor even for full context, so that cell should be treated as a prompt/model
weakness rather than a memory-strategy result.  The result is useful for the
paper's protocol claim because it shows budget enforcement, exact-match
memory scoring, paired comparisons, and category-level diagnostics.  It should
not be used as a public benchmark claim until corroborated with an external
memory-agent dataset.

\subsection{Public LongMemEval Oracle Study}

To corroborate the synthetic result on public data, we added an adapter for
the official LongMemEval JSON format~\citep{wu2024longmemeval}.  The adapter
loads \path{longmemeval_oracle.json}, formats timestamped user-assistant
sessions as a budgeted chat history, and records LongMemEval question type as
the memory category.  On June 23, 2026, we evaluated all 500 oracle-file
questions with \texttt{openai/gpt-4o-mini} through OpenRouter at temperature
0, using a 64-token output cap.  The matrix covers truncation, RAG, lean
retrieval, and full context at 2K, 4K, and 8K active-input budgets, for 6,000
item--strategy--budget rows.

The budget enforcer admitted 5,122 prompts and rejected 878 full-context
prompts before inference.  Every admitted prediction was exported to the
upstream LongMemEval hypothesis format and scored with the upstream
\path{evaluate_qa.py} evaluator using \texttt{openai/gpt-4o}; all 5,122
predictions received a parseable judge label.  The deterministic normalized-
containment metric remains in the raw BudgetBench logs as a diagnostic, but
the claim-bearing accuracies in Table~\ref{tab:longmem_oracle_pilot} are the
GPT-4o evaluator labels.  Per-cell hypotheses and evaluator outputs use the
prefix \path{results/longmem_openrouter_full_20260623_}, and the consolidated
table is \path{results/longmem_openrouter_full_20260623_official_summary.csv}.
Model, provider, row-count, and cost metadata are in
\path{results/longmem_openrouter_full_20260623_run_metadata.json}.

\begin{table}[t]
\centering
\small
\resizebox{\textwidth}{!}{%
\begin{tabular}{llrrrrrr}
\toprule
Strategy & Budget & Eval $N$ & Pred $N$ & Violations & Local contains & Official GPT-4o & Candidate sec \\
\midrule
Truncation & 2K & 500 & 500 & 0 & 0.3160 & 0.3800 & 553.8 \\
Truncation & 4K & 500 & 500 & 0 & 0.4300 & 0.5700 & 675.2 \\
Truncation & 8K & 500 & 500 & 0 & 0.4880 & 0.6500 & 952.0 \\
RAG & 2K & 500 & 500 & 0 & 0.4920 & 0.6460 & 648.5 \\
RAG & 4K & 500 & 500 & 0 & 0.5060 & 0.6860 & 839.9 \\
RAG & 8K & 500 & 500 & 0 & 0.5080 & 0.6860 & 895.0 \\
Lean retrieval & 2K & 500 & 500 & 0 & 0.4940 & 0.6640 & 593.0 \\
Lean retrieval & 4K & 500 & 500 & 0 & 0.5020 & 0.6860 & 763.4 \\
Lean retrieval & 8K & 500 & 500 & 0 & 0.4960 & 0.6820 & 944.6 \\
Full context & 2K & 500 & 48 & 452 & 0.0700 & 0.9792 & 52.1 \\
Full context & 4K & 500 & 157 & 343 & 0.2180 & 0.8854 & 190.1 \\
Full context & 8K & 500 & 417 & 83 & 0.4580 & 0.7338 & 659.7 \\
\bottomrule
\end{tabular}
}
\caption{Five-hundred-item public LongMemEval oracle-file study scored by the
upstream evaluator with \texttt{openai/gpt-4o}.  Retrieval-augmented generation
(RAG) denotes the episodic retrieval baseline.  Full-context rows are
feasibility diagnostics conditional on the budget-eligible subset shown by
Pred $N$ and must not be compared directly with the complete 500-question
strategy cells.  The run metadata records candidate inference cost
\$2.808560357, GPT-4o judging cost \$2.203379635, and total study cost
\$5.011939992.}
\label{tab:longmem_oracle_pilot}
\end{table}

The complete cells show a clear budget sensitivity for truncation, rising from
0.380 at 2K to 0.650 at 8K.  RAG and lean retrieval are substantially stronger
at 2K (0.646 and 0.664) and converge at 4K (both 0.686); their 4K and 8K Wilson
intervals overlap, so the study does not support a ranking between those two
retrieval variants.  The apparently higher full-context values are selection-
conditioned: only short/easy-enough histories survive the active-budget gate,
especially at 2K, and therefore those rows are feasibility diagnostics rather
than comparable baselines.  Candidate inference cost \$2.8086 and GPT-4o
judging cost \$2.2034 through OpenRouter, for \$5.0119 total.

This study supplies official-evaluator evidence for the oracle-file adapter,
but it is not a LongMemEval\_S full-history leaderboard result.  The oracle file
contains the evidence-focused histories used by this integration, and the
candidate model is API-hosted rather than local.  Accordingly, the result
strengthens the protocol and strategy-comparison evidence while leaving local
large-model and full-history validation as separate future work.  All released
artifacts are collected in the public repository at \RepoURL{}.

\subsection{Supplementary Model and Infrastructure-Only Multimodal Checks}

Table~\ref{tab:model_transfer_gemma4} reports the nonzero rows from the
follow-up small-model sweep, and Table~\ref{tab:model_transfer_excluded}
reports the zero or endpoint-mismatch rows from the same sweep.  This is not a
ranking comparison with the 89-item-per-task run: it uses only three LongBench
v2 items and only the 2K--8K tiers.  Its purpose is to verify that the
\BudgetBench{} protocol can be reused across newly installed Ollama tags while
keeping negative and uninformative runs visible.  On this slice,
\texttt{gemma4:e2b} produced nonzero exact-match accuracy in every reported
cell.  RAG had equal or higher measured accuracy than truncation, while
incurring higher single-run mean cell duration.

Table~\ref{tab:multimodal_probe} reports the separate native multimodal probe
as infrastructure validation only.  The six generated items use simple
image-plus-text prompts, so the result should not be read as evidence about
real-world visual reasoning or multimodal memory.  The probe verifies that the
local serving stack can carry image payloads and that \BudgetBench{} can record
multimodal artifacts without mixing them into the text-only LongBench budget
table.

\begin{table}[t]
\centering
\caption{Nonzero rows from the additional small-model LongBench v2 transfer diagnostic.  Retrieval-augmented generation (RAG) denotes the episodic retrieval baseline.  The check used three LongBench v2 items, the 2K--8K budget tiers, temperature 0.0, and Ollama's OpenAI-compatible endpoint.  These local alias rows lack complete model provenance and are not benchmark claims.  Zero-accuracy and endpoint-mismatch rows from the same sweep are reported separately in Table~\ref{tab:model_transfer_excluded}.}
\label{tab:model_transfer_gemma4}
\begin{tabular}{llcccc}
\toprule
Model & Strategy & 2K & 4K & 8K & Mean duration (s) \\
\midrule
\texttt{gemma4:e2b} & Truncation & 0.33 & 0.67 & 0.67 & 23.7 \\
\texttt{gemma4:e2b} & RAG & 0.67 & 0.67 & 0.67 & 46.6 \\
\bottomrule
\end{tabular}
\end{table}

\begin{table}[t]
\centering
\caption{Zero-accuracy and endpoint-mismatch rows from the additional small-model LongBench v2 transfer diagnostic.  Retrieval-augmented generation (RAG) denotes the episodic retrieval baseline.  All entries use three LongBench v2 items and report exact-match accuracy at 2K, 4K, and 8K.  These local alias rows lack complete model provenance and are not benchmark claims.}
\label{tab:model_transfer_excluded}
\resizebox{\textwidth}{!}{%
\begin{tabular}{llcl}
\toprule
Model & Strategy & Accuracy at 2K / 4K / 8K & Reporting note \\
\midrule
\texttt{qwen3-vl:2b} & Truncation & 0.00 / 0.00 / 0.00 & Text-only endpoint check \\
\texttt{qwen3-vl:2b} & RAG & 0.00 / 0.00 / 0.00 & Text-only endpoint check \\
\texttt{qwen3.5:0.8b} & Truncation & 0.00 / 0.00 / 0.00 & Zero exact-match signal \\
\texttt{qwen3.5:0.8b} & RAG & 0.00 / 0.00 / 0.00 & Zero exact-match signal \\
\texttt{qwen3.5:2b} & Truncation & 0.00 / 0.00 / 0.00 & Zero exact-match signal; 1,024-token output cap \\
\texttt{qwen3.5:2b} & RAG & 0.00 / 0.00 / 0.00 & Zero exact-match signal; 1,024-token output cap \\
\bottomrule
\end{tabular}
}
\end{table}

\begin{table}[t]
\centering
\caption{Native multimodal probe result for a generated six-item image question answering (QA) set.  Each item contains a simple rendered colored shape and a multiple-choice text question.  This local alias row is a modality-path diagnostic, not a benchmark-scale vision evaluation or reproducible model claim.}
\label{tab:multimodal_probe}
\begin{tabular}{llccc}
\toprule
Model & Modality & Items & Accuracy & Mean duration (s) \\
\midrule
\texttt{qwen3-vl:2b} & Image + text & 6 & 1.00 & 1.7 \\
\bottomrule
\end{tabular}
\end{table}

We also ran a 20-item full-context-feasible transfer check with
\texttt{gemma4:e2b}, using the same LongBench v2 filtering rule and the 8K and
32K active budgets.  The corresponding artifacts are tracked in \RepoURL{}.
This non-provenanced diagnostic check did not
reproduce the \RunModel{} quality signal: truncation and RAG scored 0.00 at
8K, while truncation, RAG, and full context each scored 0.05 at 32K.  Full
context at 8K again had 100\% violation rate.  The result is useful as a
transfer warning rather than a strategy ranking.  It shows that the harness
can expose model-dependent prompt/serving suitability, and that the 50-item
Qwen result should not be generalized across tasks or model scales without
checking for zero-signal runs.

We then ran \texttt{qwen3.5:2b} on a 5-item LongBench slice at 8K and 32K.
That run completed quickly enough to be practical, but every reported cell was
0.00 across truncation, RAG, lean retrieval, and full context.  This is a
useful negative transfer result: the protocol can complete on a smaller local
model without crashing, but the QA signal can collapse to zero.

Table~\ref{tab:longmem_transfer_probes} summarizes the remaining
non-reproducible local-alias transfer probes in this section.  We additionally
ran \texttt{qwen3.6:35b-mlx} on a 10-item LongMemEval oracle
slice.  Under the deterministic containment scorer, the budgeted truncation, RAG, and lean-retrieval rows all scored 1.00
at 2K, 4K, and 8K, while full context scored 0.00 at 2K, 0.20 at 4K, and 0.70
at 8K with violation rates of 1.00, 0.80, and 0.30.  The paired deltas
versus \texttt{full\_context@8192} are +0.30 for every budgeted row.  This is
an infrastructure diagnostic for a local alias rather than benchmark evidence:
the tag lacks complete source-registry, quantization-hash, and
conversion-command provenance, and the containment scorer is not the official
judge.

By contrast, a matched 10-item \texttt{gemma4:12b-mlx} LongMemEval probe
returned 0.00 across truncation, RAG, lean retrieval, and full context at
2K, 4K, and 8K.  That flat result is useful because it shows the same protocol
does not magically produce a LongMemEval signal on a smaller intermediate
model; model scale and serving regime remain first-order variables in this
evaluation.

A \texttt{checkpoint\_context} run with \RunModel{} on the same 10-item
LongMemEval slice lands at 0.10 at 2K, 0.30 at 4K, and 0.10 at 8K.  That row is
lower than the Qwen3.6 public-memory rows above, but it has a distinct shape:
the checkpoint policy behaves like a mid-budget compromise rather than a
balanced hybrid strategy.

\begin{table}[t]
\centering
\small
\setlength{\tabcolsep}{3pt}
\begin{tabular}{llll}
\toprule
Model & Strategy group & Items & Accuracy at 2K / 4K / 8K \\
\midrule
\texttt{qwen3.6:35b-mlx} & Truncation, RAG, lean retrieval & 10 & 1.00 / 1.00 / 1.00 \\
\texttt{qwen3.6:35b-mlx} & Full context & 10 & 0.00 / 0.20 / 0.70 \\
\texttt{gemma4:12b-mlx} & Truncation, RAG, lean retrieval, full context & 10 & 0.00 / 0.00 / 0.00 \\
\RunModel{} & Checkpoint context & 10 & 0.10 / 0.30 / 0.10 \\
\bottomrule
\end{tabular}
\caption{Supplementary non-reproducible local-alias LongMemEval oracle transfer diagnostics.  Retrieval-augmented generation (RAG) denotes the episodic retrieval baseline.  For the \texttt{qwen3.6:35b-mlx} full-context row, the violation rates at 2K, 4K, and 8K are 1.00, 0.80, and 0.30.  These rows are not benchmark claims without complete model provenance and official-judge validation.}
\label{tab:longmem_transfer_probes}
\end{table}

\section{Discussion}

\subsection{Quality Curves Are Not Guaranteed to Be Monotonic}

The pilot reinforces a point that long-context evaluation has repeatedly
exposed: increasing the active context budget does not necessarily improve
quality.  A larger prompt can include more evidence, but it can also include
more distractors and can change the relative position of salient information.
For a small model, the additional content may be more harmful than helpful.
This is especially visible in LongBench-style tasks, where the model must
select a single answer from a long context rather than merely continue a short
conversation.

\subsection{Compliance Failures Are First-Class Results}

Budget violations are not peripheral errors.  They tell us whether a memory
strategy can actually operate at a requested tier.  In the pilot, tight budgets
can fail when the irreducible system-plus-query prompt already exceeds the
budget after strategy fallback.  Such failures should be counted separately
from low task quality: a strategy that fails to produce a budget-compliant
prompt is operationally different from one that produces a compliant but wrong
answer.

\subsection{Latency Can Dominate Strategy Choice}

The summary-buffer baseline is conceptually attractive because it attempts to
preserve global state rather than dropping it.  In practice, it can add a full
extra generation call before the answer call.  On long-context tasks, that
extra call dominates runtime.  A memory strategy that improves accuracy by a
small amount may still be unattractive if it multiplies latency.  Conversely,
a retrieval strategy can have a fixed embedding cost but avoid repeated
generation-based compression.  This is why \BudgetBench{} reports duration and
budget utilization alongside quality.  The 50-item full-context-feasible slice
shows the reverse operational pattern: RAG at 8K has no detectable paired
accuracy difference from 32K full context at this sample size while remaining
budget-compliant at tighter tiers.  The associated wall-clock values are
single-run diagnostics, not a stable latency result.  That result is not an
equivalence result or a general strategy ranking.  It is primarily a
budget-size and feasibility tradeoff of the kind a budgeted protocol is meant
to reveal.

\subsection{Industry ROI and Adoption Path}

\BudgetBench{} is best understood as solving a measurement and reproducibility
problem for performance-sensitive deployments, not as a paper that already
selects a universal memory strategy.  Industry teams building local, edge,
air-gapped, latency-sensitive, or cost-constrained agents often need to choose
among truncation, summarization, retrieval, and hybrid memory policies under a
hard per-call context budget.  Without a shared protocol, those comparisons are
usually ad hoc: one team measures quality, another measures latency, a third
silently drops over-budget prompts, and none of the resulting numbers are
directly comparable.  The immediate return on investment is therefore
infrastructure-tier value: teams can reuse a budget enforcer, violation-rate
metric, strategy interface, prompt audit trail, and bootstrap analysis path
instead of rebuilding them for every internal agent.

This distinction affects how the artifact should be used.  The current pilot
does not justify a deployment rule such as ``use RAG instead of truncation.''
The strategy-versus-full-context direction is underpowered or contradictory
across slices.  What the artifact does provide is the machine for answering
that question inside a specific deployment envelope.  A team can pin its model,
hardware, target budget, task subset, and service objective, then ask whether a
candidate strategy improves quality without increasing violation rate or
latency beyond the product tolerance.

For adoption, we recommend four concrete steps.  First, define one or two
target operating tiers, such as 4K for an interactive local assistant or 8K for
an air-gapped analyst workflow, before running a full sweep.  Second, treat
budget-violation rate as a release gate rather than a secondary diagnostic:
an over-budget strategy is operationally different from a compliant but wrong
answer.  Third, run at least 100 task items per claim-bearing comparison, with
paired item-level estimates and repeated cells for latency-sensitive decisions.
Fourth, require every candidate memory policy to declare its deployment regime:
local-only, auxiliary-model-assisted, hosted, persistent-stateful, or
reset-per-item.  Those declarations are part of the cost model, not metadata.

\subsection{Pilot Scope}

The current runs are intentionally small.  They exercise the measurement
surface, not final strategy rankings.  The 50-item
feasible-slice result is enough to justify scaling because it includes a
paired full-context comparison and shows a compliance tradeoff, but it does
not settle the direction of the budgeted-versus-full-context comparison.  The
hosted Qwen3 30B-A3B replication shows that the same protocol can run with
exact tokenization and repeated shuffled cells on a stronger model, but its
point estimate favors full context and it does not establish local-serving
transfer.
A benchmark-scale study should still use stronger fully specified local models,
extend the new 500-question LongMemEval oracle study to LongMemEval\_S or other
full-history settings, add official SWE-bench resolution or a deterministic trace-debugging
alternative, include richer strategy families, and add multimodal task
families before making claims about vision-language memory.  The pilot
establishes the path from local serving to budget-constrained logs, analysis,
tables, figures, and a reproducible paper artifact.

\section{Threats to Validity}

\paragraph{Internal validity.}
The runner uses a tokenizer function based on \texttt{tiktoken} rather than
the exact Ollama model tokenizer.  A post-hoc audit of the first 89 LongBench
v2 natural prompts with the \texttt{Qwen/Qwen2.5-1.5B-Instruct} tokenizer found
that Qwen counts average 5.5\% higher than \texttt{cl100k\_base} counts
(median 4.1\%; range -6.4\% to +31.4\%).  Natural full-context feasibility is
unchanged at 2K, 4K, 8K, and 32K in this audit, with one 16K item flipping from
fit under \texttt{tiktoken} to over-budget under Qwen tokenization.  For
reconstructed LongBench truncation prompts selected with the \texttt{tiktoken}
budget enforcer, Qwen recounting would mark 14/89, 35/89, 53/89, 65/89, and
50/89 prompts over budget at 2K, 4K, 8K, 16K, and 32K, respectively.  Thus the
current violation-rate rows should be read as compliance with the benchmark
tokenizer approximation, not compliance with the served model tokenizer.  This
is not a full strategy-matrix recount because the historical pilot logs do not store
strategy-processed prompts for retrieval or summary contents.  The current
runner has since been upgraded to log tokenizer identifiers, prompt hashes,
post-strategy prompt artifacts, and explicit token counts.  The 89-item
LongBench tables in this paper therefore remain exploratory
tokenizer-approximation diagnostics rather than claim-bearing compliance
tables.  The hosted stronger-model table is an exception: it uses the exact Qwen3 tokenizer for selection,
strategy processing, enforcement, and logged prompt audits.  A newer 10-item LongBench
    rerun in the repository artifact bundle demonstrates the
corrected harness path with explicit Qwen tokenization and repeated
order-randomized trials, but it is still too small to replace the older
exploratory matrix.  The
summary strategy uses the same small model for compression; a stronger
summarizer could change the results.  The RAG strategy uses generic embeddings
and simple message-level retrieval rather than task-optimized chunking.
Most reported local latency values are single-run wall-clock measurements.  They were
not produced with repeated warmup-controlled trials or randomized cell order,
so they should be read as operational diagnostics rather than stable latency
benchmarks.  The hosted stronger-model cells use two shuffled repeats and
report across-repeat standard deviations, but two trials remain insufficient
for a stable provider-latency estimate and include network and queue time.
LLM serving latency also depends on request scheduling and
key-value cache management, especially for long prompts~\citep{kwon2023pagedattention}.

\paragraph{External validity.}
The model is small, the primary sample contains 89 items per task, and the
memory-agent task is synthetic.  Larger local models may benefit more from
larger budgets.  Production agents may maintain cross-item or cross-session
memory, whereas this pilot resets strategy state between items.  The results
should therefore be interpreted as a validation of the protocol and
implementation, not as a general statement about all local memory strategies.
Per-item resets also exclude persistent index construction, maintenance, and
cross-session retrieval costs that can dominate real deployments.  Production
cost comparisons should therefore add a persistent-memory regime rather than
using the reset-per-item pilot timings alone.
The additional \texttt{gemma4:e2b} sweep is also narrow: it is a 20-item
LongBench transfer check and should be read only as evidence that the protocol
can expose model-dependent prompt and serving behavior.
The stronger LongBench replication uses an API-hosted open-weight model.  It
reduces the small-model external-validity concern for task quality, but does
not validate local memory pressure, key-value-cache behavior, energy use, or
latency on a 30B-class local serving stack.

\paragraph{Construct validity.}
The SWE patch-similarity proxy measures overlap with the reference patch, not
actual repository repair.  It can reward patches that touch the right file or
line without being executable fixes.  LongBench v2 exact-match scoring is more
direct, but 89 questions are still too few for definitive estimates.  The
LongMemEval oracle study uses the upstream evaluator with GPT-4o, eliminating
the earlier normalized-containment grading proxy for its main accuracy table.
It remains an oracle-file rather than LongMemEval\_S full-history evaluation,
and GPT-4o judgment is itself a model-based construct that can vary across
judge versions.  Full-context rows are additionally conditioned on passing
the budget gate and therefore measure a selected subset.  We report Wilson
intervals and avoid comparing those selected rows with complete 500-question
cells.

\paragraph{Multimodal validity.}
The main text-task pilot does not contain image, video, chart, or
document-page inputs.  The added image QA probe verifies the native
image-plus-text path on generated colored shapes, but it is too small and too
simple to support claims about real multimodal memory.  Therefore, this paper
should be read as adding a multimodal data path and extension plan, not as
claiming benchmark-scale multimodal performance.

\section{Planned Extensions}

The next version of \BudgetBench{} should expand in six directions.  First,
it should replace proxy SWE grading with official containerized evaluation.
Second, it should include $\tau$-bench or tau2-bench once the local data
dependency is installed, because tool-use chains stress memory in ways that
single-call QA does not.  Third, it should add richer strategy families:
LLMLingua-2 compression, Mem0, Letta/MemGPT-style hierarchical memory, and
hybrid retrieval-summary policies.  Fourth, it should report the analyzer's
confidence intervals in every main quality table and use the now-supported
repeated-cell execution path so strategy crossovers can be
distinguished from sampling noise beyond the current two-repeat hosted study.
Fifth, it should add a persistent-memory regime that reports index build time,
index maintenance cost, and amortized index cost per evaluated item.  Sixth,
it should add multimodal tasks with
native visual inputs and multimodal memory policies.  Examples include
document-page QA where the active
budget controls optical character recognition (OCR) text plus image references, chart QA where retrieval
chooses both visual crops and textual evidence, and tool-agent tasks where
screenshots or UI state are part of the memory trace.

\section{Conclusion}

\BudgetBench{} operationalizes an under-standardized evaluation
question: how do memory strategies trade quality, budget compliance, and
latency when active context is fixed?  The expanded 89-item-per-task pilot run
and the 50-item full-context-feasible LongBench slice show that this question
can be evaluated locally with a small model, deterministic graders, fixed
budget tiers, and swappable strategy baselines.  The strongest claim is
methodological rather than rank-oriented: fixed-budget evaluation reveals
failure modes, infeasible cells, tokenizer-accounting gaps, and operating
points that standard long-context evaluation can hide.  The empirical
budgeted-versus-full-context direction is unresolved in the present artifact:
the 50-item local slice is near-null, the 10-item tokenizer-aligned hardened
slice is too small to rank strategies, and the hosted stronger-model slice
favors full context in point estimate.  The synthetic pilot and 500-question
LongMemEval oracle study show that the same protocol can expose
memory-specific update, preference, temporal, and abstention behavior, but the
well-powered runs are hosted and the local runs remain pilot-scale.  The
reusable contribution is the active-budget protocol and harness, together with
the reproducibility and failure-reporting discipline needed to scale it into a
community benchmark.

\clearpage
\appendix

\section{Reproducibility Checklist}

Table~\ref{tab:repro_checklist} separates claim-bearing reproducible artifacts
from non-reproducible diagnostic artifacts.  Local aliases without complete
source-registry, quantization-hash, and conversion-command provenance are kept
for auditability but are not benchmark claims.

\begin{longtable}{L{0.22\textwidth}L{0.68\textwidth}}
\caption{Reproducibility checklist split by claim status.  QA denotes question answering, RAG denotes retrieval-augmented generation, CSV denotes comma-separated values, and PNG denotes portable network graphics.}\label{tab:repro_checklist}\\
\toprule
Item & Value \\
\midrule
\endfirsthead
\toprule
Item & Value \\
\midrule
\endhead
\multicolumn{2}{l}{\textbf{Reproducible / claim-bearing or claim-supporting artifacts}} \\
Run date & June 16, 2026 \\
Model & \RunModel{} served through Ollama \\
Endpoint & \OllamaEndpoint{} \\
Run identifier & \RunId{} \\
Budget tiers & 2,048; 4,096; 8,192; 16,384; 32,768 active input tokens \\
Tasks & SWE-bench Verified scaffold; LongBench v2 multiple-choice QA \\
Strategies & Truncation, summary buffer, episodic RAG \\
Items per task & 89 \\
Temperature & 0.0 \\
Maximum generated tokens & 512 \\
Random seed & 42 \\
Primary pilot artifacts & \RepoURL{} \\
Primary aggregate CSV & \path{results/full_study_qwen2_5_1_5b_89items.csv} \\
Full-context LongBench slice & Repository artifact bundle; aggregate \path{results/full_study_qwen2_5_1_5b_20260618_155102.csv} \\
Tokenizer-aligned LongBench hardened slice & Repository artifact bundle; aggregate \path{results/full_study_qwen2_5_1_5b_20260623_112431.csv}; paired \path{results/paired_deltas_qwen2_5_1_5b_20260623_112431.csv} \\
Hosted stronger-model LongBench replication & 50 items; two shuffled repeats; \texttt{qwen/qwen3-30b-a3b-instruct-2507} through OpenRouter; exact \texttt{Qwen/Qwen3-30B-A3B-Instruct-2507} tokenizer; aggregate \path{results/full_study_qwen_qwen3-30b-a3b-instruct-2507_20260624_123733.csv}; paired \path{results/paired_deltas_qwen_qwen3-30b-a3b-instruct-2507_20260624_123733.csv}; metadata \path{results/longbench_qwen3_30b_openrouter_50x2_20260624_run_metadata.json} \\
RAG ablation slice & Repository artifact bundle; comparison \path{results/ablation_rag_long10_20260623.csv} \\
Official SWE harness rerun & \path{qwen2.5:1.5b.budgetbench_official2fix_20260623.json}; tracked in the repository artifact bundle \\
Synthetic memory pilot & Repository artifact bundle; aggregate \path{results/full_study_qwen2_5_1_5b_20260618_203210.csv}; categories \path{results/grouped_memory_category_qwen2_5_1_5b_20260618_203210.csv} \\
LongMemEval oracle study & 500 items; \texttt{openai/gpt-4o-mini} candidate through OpenRouter; \texttt{openai/gpt-4o} upstream evaluator; summary \path{results/longmem_openrouter_full_20260623_official_summary.csv}; run metadata \path{results/longmem_openrouter_full_20260623_run_metadata.json}; per-cell hypotheses and labels under the same results prefix; repository artifact bundle \\
\multicolumn{2}{l}{\textbf{Non-reproducible diagnostics / provenance incomplete}} \\
Qwen3.6-35B LongMemEval probe & Local alias diagnostic only; aggregate \path{results/full_study_qwen3_6_35b-mlx_20260618_212102.csv}; paired \path{results/paired_deltas_qwen3_6_35b-mlx_20260618_212102.csv} \\
Qwen3.6-35B LongMemEval transfer & Local alias diagnostic only; aggregate \path{results/full_study_qwen3_6_35b-mlx_20260619_140108.csv}; paired \path{results/paired_deltas_qwen3_6_35b-mlx_20260619_140108.csv}; categories \path{results/grouped_memory_category_qwen3_6_35b-mlx_20260619_140108.csv} \\
Qwen3.5-2B LongBench 5-item transfer & Local alias diagnostic only; aggregate \path{results/full_study_qwen3_5_2b_20260619_142423.csv}; paired \path{results/paired_deltas_qwen3_5_2b_20260619_142423.csv} \\
Gemma transfer check & Local alias diagnostic only; aggregate \path{results/full_study_gemma4_e2b_20260618_161005.csv} \\
Additional filtered model sweep & Diagnostic only; \path{results/model_compare_small_long3.csv}; nonzero and zero-accuracy rows are both reported in the supplementary model tables \\
Supplementary local tag caveat & \texttt{gemma4:e2b}, \texttt{qwen3.5:0.8b}, \texttt{qwen3.5:2b}, \texttt{qwen3-vl:2b}, \texttt{qwen3.6:35b-mlx}, \texttt{gemma4:12b-mlx}, \texttt{gemma4:31b-mlx}, and \texttt{gemma3:1b} are local aliases without complete provenance records; their rows are not benchmark claims \\
Native multimodal probe artifacts & \path{results/multimodal_smoke_qwen3_vl_2b.csv}; generated PNGs in \path{results/multimodal_smoke/} \\
\bottomrule
\end{longtable}

\section{Task Specifications}

Tables~\ref{tab:task_swe_card},~\ref{tab:task_longbench_card},~\ref{tab:task_synth_memory_card}, and~\ref{tab:task_longmemeval_card}
define the task cards used by the pilot and follow-up evaluations.

\subsection{SWE-bench Verified Scaffold}

\begin{longtable}{L{0.24\textwidth}L{0.66\textwidth}}
\caption{SWE scaffold task card.}\label{tab:task_swe_card}\\
\toprule
Field & Description \\
\midrule
\endfirsthead
\toprule
Field & Description \\
\midrule
\endhead
Purpose & Evaluate whether a local model can transform a natural-language bug report and hints into a patch-like diff under constrained context. \\
Input format & System prompt instructing the model to produce only a unified diff; user prompt containing the problem statement and available hints. \\
Prediction format & Text response from which the first \texttt{diff --git} block is extracted. \\
Grader & Deterministic patch-similarity proxy using file recall and changed-line overlap. \\
Strength & Low-cost, deterministic, and fast enough for repeated budget sweeps. \\
Limitation & Does not execute tests; cannot be interpreted as official SWE-bench resolution. \\
Follow-up artifact & A focused official harness rerun is included in \path{qwen2.5:1.5b.budgetbench_official2fix_20260623.json}; one invalid patch is filtered before official evaluation, and the remaining two submitted patches still fail official patch application. \\
Scaling path & Replace the proxy table with larger official SWE-bench reruns and stratify by repository, patch size, and issue type. \\
\bottomrule
\end{longtable}

\subsection{LongBench v2}

\begin{longtable}{L{0.24\textwidth}L{0.66\textwidth}}
\caption{LongBench v2 task card.}\label{tab:task_longbench_card}\\
\toprule
Field & Description \\
\midrule
\endfirsthead
\toprule
Field & Description \\
\midrule
\endhead
Purpose & Evaluate long-context evidence selection and answer-choice discrimination under fixed active budgets. \\
Input format & System prompt, context chunks, question, and four answer choices. \\
Prediction format & One of A, B, C, or D, extracted from the model response. \\
Grader & Exact-match letter comparison. \\
Strength & Deterministic and directly compatible with budgeted prompt manipulation. \\
Limitation & Small pilot subset; exact-match increments are coarse at low sample counts. \\
Scaling path & Increase sample count; stratify by context length, evidence dispersion, and reasoning type. \\
\bottomrule
\end{longtable}

\subsection{Synthetic Memory Pilot}

\begin{longtable}{L{0.24\textwidth}L{0.66\textwidth}}
\caption{Synthetic memory pilot task card.}\label{tab:task_synth_memory_card}\\
\toprule
Field & Description \\
\midrule
\endfirsthead
\toprule
Field & Description \\
\midrule
\endhead
Purpose & Evaluate whether a strategy can preserve or retrieve memory-relevant events under tight active budgets. \\
Input format & System prompt, event history with distractors, and a direct question requiring a short exact answer. \\
Categories & Single-session fact, knowledge update, temporal reasoning, preference, and abstention. \\
Prediction format & Short text answer normalized for exact-match scoring. \\
Grader & Deterministic normalized exact match. \\
Strength & Fast, reproducible, and memory-specific without requiring an LLM judge. \\
Limitation & Synthetic; not a substitute for public memory-agent benchmarks such as LongMemEval or LoCoMo. \\
Scaling path & Replace or corroborate with public memory datasets; stratify by category and distractor density. \\
\bottomrule
\end{longtable}

\subsection{LongMemEval Oracle Study}

\begin{longtable}{L{0.24\textwidth}L{0.66\textwidth}}
\caption{LongMemEval oracle study task card.}\label{tab:task_longmemeval_card}\\
\toprule
Field & Description \\
\midrule
\endfirsthead
\toprule
Field & Description \\
\midrule
\endhead
Purpose & Evaluate public long-term chat memory data under BudgetBench active-context tiers. \\
Input format & Timestamped LongMemEval oracle sessions flattened into chat messages, followed by the dated question. \\
Categories & LongMemEval question type, including abstention, knowledge update, multi-session, single-session, and temporal reasoning variants. \\
Prediction format & Short free-form text answer. \\
Grader & Upstream LongMemEval \path{evaluate_qa.py} with \texttt{openai/gpt-4o}; normalized containment is retained only as a local diagnostic. \\
Strength & Public memory-agent data with natural long histories and category metadata. \\
Limitation & Five hundred oracle-file questions; not a LongMemEval\_S full-history run.  Full-context accuracy is conditional on budget eligibility, and candidate inference is API-hosted rather than local. \\
Scaling path & Run LongMemEval\_S/full-history variants with stronger fully specified local models and repeat judge-version sensitivity checks. \\
\bottomrule
\end{longtable}

\section{Strategy Specifications}

\subsection{Full Context}

The full-context baseline returns the task formatter's natural message list
unchanged.  It is not a memory policy and is not included in the 89-item-per-task pilot
tables.  Its purpose is to support follow-up comparisons against the largest
prompt that the model/server can accept and to make infeasible full-context
cells visible as budget violations rather than silent exclusions.

\subsection{Truncation}

Truncation is the simplest baseline.  It preserves the system prompt and then
walks backward through the message list, retaining the newest messages that fit
inside the remaining budget.  Its primary failure mode is the loss of older
evidence, including evidence that may be decisive.  Its advantage is that it
requires no extra model calls, no embedding model, and no external memory
store.  In a budget benchmark, truncation is the minimal memory-management
baseline that more complex strategies should be compared against.

\subsection{Summary Buffer}

The summary-buffer strategy attempts to compress evicted context into a
shorter natural-language state.  The pilot implementation reserves a portion
of the budget for a summary, chooses recent messages to retain, and asks the
local model to summarize the evicted messages.  This baseline can preserve
global state, but it introduces additional generation cost and can suffer from
summary drift.  In the pilot, summary is especially expensive on LongBench
because long contexts trigger repeated compression calls.

\subsection{Episodic RAG}

The RAG baseline indexes intermediate messages and retrieves chunks relevant
to the current query.  It is a direct adaptation of retrieval-augmented
generation to the prompt-budget setting~\citep{lewis2020rag}.  Its advantage
is selectivity: it can recover older evidence without including the full
history.  Its limitations are query dependence, chunking sensitivity, embedding
cost, and the possibility that relevant information is not semantically close
to the final question.

\subsection{Lean Retrieval}

The lean retrieval baseline combines dense similarity, lexical overlap,
recency, and lightweight salience signals before packing selected messages
into the active budget.  It is designed as a local, no-database contrast to
the episodic RAG implementation.  In the current evidence it is a mixed
baseline: it does not beat simple RAG on the LongBench feasible slice, but it
is the strongest row in the synthetic memory pilot.  This is useful for
diagnostics because it suggests that strategy crossovers may depend on
whether evidence is semantically localized, temporally updated, or dispersed
among distractors.

\section{Metric Definitions}

\subsection{Quality}

Quality is task-specific.  For LongBench v2 it is exact-match accuracy:
\[
  \mathrm{Acc} = \frac{1}{N}\sum_i \mathbb{1}[\hat{a}_i = a_i].
\]
For the SWE scaffold it is a continuous patch-similarity score:
\[
  \mathrm{SWEProxy} = 0.4 \cdot \mathrm{FileRecall} + 0.6 \cdot \mathrm{LineOverlap}.
\]
The proxy is useful for pilot measurement but should be replaced by official
test execution for benchmark claims.
For the synthetic memory pilot, quality is normalized exact-match accuracy on
short answers.

\subsection{Violation Rate}

Violation rate is the mean retry-level budget failure rate over items in a
cell.  If a strategy returns a prompt exceeding the active tier, the runner
logs the violation and retries.  A cell can therefore have nonzero quality and
nonzero violations if some items succeed while others fail under the tier.

\subsection{Utilization and Peak Budget}

Mean used budget measures how many prompt tokens the strategy actually sends
on successful calls.  Maximum peak budget records the largest token count
observed during enforcement, including failed attempts.  These metrics separate
conservative strategies that under-use budget from strategies that operate
near the ceiling.

\subsection{Duration}

Duration is wall-clock time for the cell.  It includes strategy overhead,
embedding or summarization calls, final model calls, and dataset-item loop
overhead.  Duration is machine-dependent, but it remains operationally useful
because \BudgetBench{} is explicitly about local deployment regimes.

\section{Full Operational Tables}

\begin{figure}[p]
  \centering
  \includegraphics[width=\textwidth]{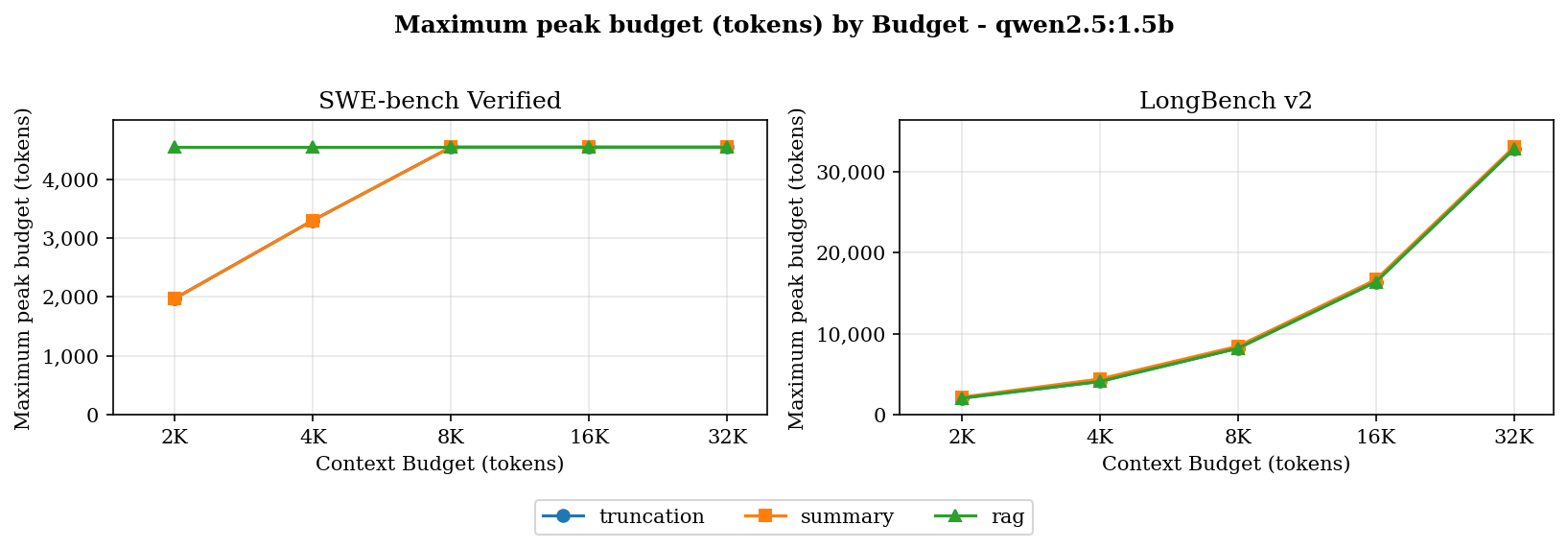}
  \caption{Maximum peak budget observed in each cell. Blue circles denote truncation, orange squares denote summary, and green triangles denote retrieval-augmented generation (RAG).}
  \label{fig:peak_budget_89item}
\end{figure}

Tables~\ref{tab:app_quality},~\ref{tab:app_violation},~\ref{tab:app_mean_used},~\ref{tab:app_peak}, and~\ref{tab:app_duration}
provide the full quality, violation, budget-use, peak-budget, and duration
matrices for the 89-item-per-task run.  Table~\ref{tab:app_flat_audit}
provides the flat per-cell aggregate audit view.  Figure~\ref{fig:peak_budget_89item}
plots the peak-budget curves from the same run.

\begin{table}[p]
\centering
\caption{Full quality matrix for the 89-item-per-task run. Retrieval-augmented generation (RAG) denotes the episodic retrieval baseline. LongBench v2 values are exact-match accuracy. SWE values are patch-similarity proxy scores and are not official SWE-bench accuracy.}
\label{tab:app_quality}
\begin{tabular}{llccccc}
\toprule
Task & Strategy & 2K & 4K & 8K & 16K & 32K \\
\midrule
\multirow{3}{*}{SWE} & Truncation & 0.18 & 0.19 & 0.19 & 0.19 & 0.19 \\
& Summary & 0.18 & 0.19 & 0.19 & 0.19 & 0.19 \\
& RAG & 0.18 & 0.19 & 0.19 & 0.19 & 0.19 \\
\midrule
\multirow{3}{*}{LongBench v2} & Truncation & 0.29 & 0.27 & 0.27 & 0.33 & 0.33 \\
& Summary & 0.30 & 0.24 & 0.26 & 0.29 & 0.31 \\
& RAG & 0.30 & 0.31 & 0.35 & 0.34 & 0.31 \\
\bottomrule
\end{tabular}
\end{table}

\begin{table}[p]
\centering
\caption{Full budget-violation matrix for the 89-item-per-task run. Retrieval-augmented generation (RAG) denotes the episodic retrieval baseline. These violation rates are benchmark-tokenizer-approximation diagnostics, not served-model-tokenizer compliance claims.}
\label{tab:app_violation}
\begin{tabular}{llccccc}
\toprule
Task & Strategy & 2K & 4K & 8K & 16K & 32K \\
\midrule
\multirow{3}{*}{SWE} & Truncation & 0.00 & 0.00 & 0.00 & 0.00 & 0.00 \\
& Summary & 0.00 & 0.00 & 0.00 & 0.00 & 0.00 \\
& RAG & 0.07 & 0.03 & 0.00 & 0.00 & 0.00 \\
\midrule
\multirow{3}{*}{LongBench v2} & Truncation & 0.00 & 0.00 & 0.00 & 0.00 & 0.00 \\
& Summary & 0.03 & 0.07 & 0.05 & 0.07 & 0.03 \\
& RAG & 0.00 & 0.00 & 0.00 & 0.00 & 0.00 \\
\bottomrule
\end{tabular}
\end{table}

\begin{table}[p]
\centering
\caption{Mean used active budget in benchmark-tokenizer-approximation tokens. Retrieval-augmented generation (RAG) denotes the episodic retrieval baseline.}
\label{tab:app_mean_used}
\begin{tabular}{llccccc}
\toprule
Task & Strategy & 2K & 4K & 8K & 16K & 32K \\
\midrule
\multirow{3}{*}{SWE} & Truncation & 561 & 646 & 788 & 788 & 788 \\
& Summary & 570 & 650 & 788 & 788 & 788 \\
& RAG & 557 & 644 & 788 & 788 & 788 \\
\midrule
\multirow{3}{*}{LongBench v2} & Truncation & 1,797 & 3,803 & 7,937 & 16,012 & 29,411 \\
& Summary & 1,584 & 3,412 & 7,489 & 15,111 & 28,899 \\
& RAG & 1,919 & 3,961 & 8,052 & 16,088 & 29,483 \\
\bottomrule
\end{tabular}
\end{table}

\begin{table}[p]
\centering
\caption{Maximum observed peak budget in benchmark-tokenizer-approximation tokens. Retrieval-augmented generation (RAG) denotes the episodic retrieval baseline.}
\label{tab:app_peak}
\begin{tabular}{llccccc}
\toprule
Task & Strategy & 2K & 4K & 8K & 16K & 32K \\
\midrule
\multirow{3}{*}{SWE} & Truncation & 1,973 & 3,299 & 4,542 & 4,542 & 4,542 \\
& Summary & 1,973 & 3,299 & 4,542 & 4,542 & 4,542 \\
& RAG & 4,542 & 4,542 & 4,542 & 4,542 & 4,542 \\
\midrule
\multirow{3}{*}{LongBench v2} & Truncation & 2,028 & 4,092 & 8,180 & 16,382 & 32,762 \\
& Summary & 2,153 & 4,398 & 8,439 & 16,658 & 33,011 \\
& RAG & 2,048 & 4,094 & 8,192 & 16,384 & 32,768 \\
\bottomrule
\end{tabular}
\end{table}

\begin{table}[p]
\centering
\caption{Cell duration in seconds. Retrieval-augmented generation (RAG) denotes the episodic retrieval baseline.}
\label{tab:app_duration}
\begin{tabular}{llccccc}
\toprule
Task & Strategy & 2K & 4K & 8K & 16K & 32K \\
\midrule
\multirow{3}{*}{SWE} & Truncation & 206.4 & 210.2 & 209.0 & 203.0 & 208.7 \\
& Summary & 219.4 & 214.8 & 218.5 & 214.9 & 207.2 \\
& RAG & 187.0 & 194.2 & 203.3 & 202.9 & 202.4 \\
\midrule
\multirow{3}{*}{LongBench v2} & Truncation & 43.3 & 84.3 & 209.4 & 637.2 & 1794.8 \\
& Summary & 1862.3 & 1982.7 & 1966.8 & 2261.3 & 3240.0 \\
& RAG & 495.5 & 539.3 & 671.1 & 1103.5 & 2250.0 \\
\bottomrule
\end{tabular}
\end{table}

\begin{longtable}{lllrrrrr}
\caption{Flat per-cell aggregate data for audit.  RAG denotes retrieval-augmented generation, and Viol. denotes violation rate.}\label{tab:app_flat_audit}\\
\toprule
Model & Task & Strategy & Budget & Quality & Viol. & Mean Used & Duration (s) \\
\midrule
\endfirsthead
\toprule
Model & Task & Strategy & Budget & Quality & Viol. & Mean Used & Duration (s) \\
\midrule
\endhead
qwen2.5:1.5b & long & rag & 2,048 & 0.303 & 0.000 & 1919.2 & 495.5 \\
qwen2.5:1.5b & long & rag & 4,096 & 0.315 & 0.000 & 3960.8 & 539.3 \\
qwen2.5:1.5b & long & rag & 8,192 & 0.348 & 0.000 & 8052.4 & 671.1 \\
qwen2.5:1.5b & long & rag & 16,384 & 0.337 & 0.000 & 16087.7 & 1103.5 \\
qwen2.5:1.5b & long & rag & 32,768 & 0.315 & 0.000 & 29482.7 & 2250.0 \\
qwen2.5:1.5b & long & summary & 2,048 & 0.303 & 0.028 & 1584.4 & 1862.3 \\
qwen2.5:1.5b & long & summary & 4,096 & 0.236 & 0.068 & 3412.5 & 1982.7 \\
qwen2.5:1.5b & long & summary & 8,192 & 0.258 & 0.045 & 7488.5 & 1966.8 \\
qwen2.5:1.5b & long & summary & 16,384 & 0.292 & 0.068 & 15110.9 & 2261.3 \\
qwen2.5:1.5b & long & summary & 32,768 & 0.315 & 0.028 & 28898.8 & 3240.0 \\
qwen2.5:1.5b & long & truncation & 2,048 & 0.292 & 0.000 & 1796.6 & 43.3 \\
qwen2.5:1.5b & long & truncation & 4,096 & 0.270 & 0.000 & 3803.2 & 84.3 \\
qwen2.5:1.5b & long & truncation & 8,192 & 0.270 & 0.000 & 7936.5 & 209.4 \\
qwen2.5:1.5b & long & truncation & 16,384 & 0.326 & 0.000 & 16011.7 & 637.2 \\
qwen2.5:1.5b & long & truncation & 32,768 & 0.326 & 0.000 & 29410.9 & 1794.8 \\
qwen2.5:1.5b & swe & rag & 2,048 & 0.177 & 0.067 & 556.8 & 187.0 \\
qwen2.5:1.5b & swe & rag & 4,096 & 0.186 & 0.034 & 644.1 & 194.2 \\
qwen2.5:1.5b & swe & rag & 8,192 & 0.191 & 0.000 & 788.3 & 203.3 \\
qwen2.5:1.5b & swe & rag & 16,384 & 0.191 & 0.000 & 788.3 & 202.9 \\
qwen2.5:1.5b & swe & rag & 32,768 & 0.191 & 0.000 & 788.3 & 202.4 \\
qwen2.5:1.5b & swe & summary & 2,048 & 0.177 & 0.000 & 570.2 & 219.4 \\
qwen2.5:1.5b & swe & summary & 4,096 & 0.186 & 0.000 & 650.5 & 214.8 \\
qwen2.5:1.5b & swe & summary & 8,192 & 0.191 & 0.000 & 788.3 & 218.5 \\
qwen2.5:1.5b & swe & summary & 16,384 & 0.191 & 0.000 & 788.3 & 214.9 \\
qwen2.5:1.5b & swe & summary & 32,768 & 0.191 & 0.000 & 788.3 & 207.2 \\
qwen2.5:1.5b & swe & truncation & 2,048 & 0.177 & 0.000 & 561.0 & 206.4 \\
qwen2.5:1.5b & swe & truncation & 4,096 & 0.186 & 0.000 & 646.2 & 210.2 \\
qwen2.5:1.5b & swe & truncation & 8,192 & 0.191 & 0.000 & 788.3 & 209.0 \\
qwen2.5:1.5b & swe & truncation & 16,384 & 0.191 & 0.000 & 788.3 & 203.0 \\
qwen2.5:1.5b & swe & truncation & 32,768 & 0.191 & 0.000 & 788.3 & 208.7 \\
\bottomrule
\end{longtable}

\section{Artifact and Reporting Policy}

\subsection{Artifact Boundaries}

The artifact is a benchmark scaffold, not a static dataset release.  The code
defines a protocol and adapters for existing datasets.  The repository is the
reproducibility bundle for this paper.  Future benchmark submissions should
preserve model hashes, dependency versions, and hardware metadata so that
budget-compliance failures can be audited.
The current repository now also includes official-format export wrappers for
both LongMemEval hypotheses and SWE-bench prediction JSONL files, but only the
LongMemEval path can be exercised directly on the existing prediction-bearing
pilot logs.  The older SWE pilot logs in this paper do not persist model patch
outputs, so official SWE export requires a rerun under the current logger.
That rerun path has now been executed through the official SWE-bench harness on
a fresh three-item slice:
\path{scripts/export_swebench_predictions.py} exports valid official-format
prediction bundles from current logs, and the official SWE-bench harness
wrapper runs once the upstream \texttt{swebench} package is installed.  The
resulting official report is
\path{qwen2.5:1.5b.budgetbench_official2fix_20260623.json}, with per-instance
artifacts in the repository bundle.  A
first three-prediction smoke rerun established the harness path, after which a
second focused rerun filtered out one structurally invalid patch before
submission.  The remaining two submitted predictions still failed during
official patch application, so this should be read as completed official
evaluation infrastructure plus a negative patch quality result, not as SWE task
success.  The LongMemEval path has now also been exercised end-to-end: 5,122
prediction-bearing rows from the 500-question oracle-file matrix were exported
and labeled by the upstream evaluator with \texttt{openai/gpt-4o}.  The
per-cell hypothesis and evaluator-result files are committed under
\path{results/longmem_openrouter_full_20260623_}.

\subsection{External Judge Bridge Smoke Tests}
\label{app:judge_bridge}

The JSON-rationale bridge in \path{scripts/run_longmem_external_judge.py} was
used for early serializer and endpoint smoke tests.  Those local judge samples
remain non-claim-bearing because several small local judges produced malformed
or unreliable decisions.  The final study instead uses the canonical path:
\path{scripts/export_longmem_official_hypotheses.py} writes upstream-format
hypotheses, and \path{scripts/run_longmemeval_official_eval.py} invokes the
upstream evaluator through OpenRouter.  A two-row smoke preceded the paid run;
the full evaluation then produced 5,122 labels with no missing or unparsable
rows.  The wrapper records the evaluator model in every result row and resolves
all input paths before changing into the upstream evaluator directory.

\subsection{Submission Policy}

A public \BudgetBench{} benchmark should require strategy implementations to
declare all external dependencies, whether they call additional LLMs, whether
they use persistent state across items, and whether they require remote APIs.
Strategies that use remote proprietary services should be reported separately
from local-only strategies because they change the deployment regime being
measured.

\subsection{Recommended Reporting Template}

Every benchmark report should include: model name and quantization, serving
stack, tokenizer used for enforcement, budget tiers, task subset, number of
items, random seed, decoding parameters, strategy names, quality curves,
violation curves, mean used budget, peak budget, latency, and raw log location.
Without these fields, it is difficult to distinguish a genuine memory-strategy
effect from a serving or measurement artifact.

\section{Additional Limitations}

This paper makes several conservative claims.  First, it does not claim
novelty on long-context evaluation alone; RULER, HELMET, LongBench,
and LongBench v2 already address long-context measurement from different
angles~\citep{bai2023longbench,bai2024longbenchv2,hsieh2024ruler,yen2024helmet}.
Second, it does not claim that retrieval, summary, or truncation is globally
best.  The pilot sample is too small.  Third, it does not claim official
SWE-bench results.  The SWE column is clearly labeled as a scaffold proxy.
The novelty is the conjunction of fixed active budget, swappable memory
strategy, local model serving, deterministic logging, and task-level quality
measurement.

\section{Statistical Scaling Plan}

The pilot tables in this paper are intentionally small, but the benchmark
protocol is designed to scale to statistically meaningful comparisons.  This
appendix section specifies how a larger \BudgetBench{} study should move from
artifact validation to inference.  The most important principle is that every
claim about strategy dominance must be tied to a task, model, budget tier, and
uncertainty estimate.  Statements such as ``RAG is best'' are underspecified.
The benchmark should instead support statements such as ``for model $M$ on
LongBench v2 items with contexts in the 8K--32K range, RAG has higher mean
exact-match accuracy than truncation at the 2K active budget with a bootstrap
95\% confidence interval that excludes zero.''
Equivalence statements require a different estimand: the authors should
pre-specify a practical equivalence margin and use an equivalence test rather
than treating a confidence interval that crosses zero as a match, following
standard two-one-sided-test guidance~\citep{lakens2017equivalence}.

\subsection{Recommended Estimands}

For a fixed model $M$, task family $T$, strategy $s$, and budget $B$, the
primary estimand is expected task quality $Q(s, B, T, M)$.  The most useful
comparative estimand is the paired difference between two strategies on the
same items:
\[
  \Delta(s_a, s_b, B, T, M) =
  \frac{1}{N}\sum_{i=1}^{N}
  \left[g(M(s_a(H_i, B)), x_i) - g(M(s_b(H_i, B)), x_i)\right].
\]
Pairing is important because task difficulty can dominate strategy effects.
The same item may be easy for every strategy or impossible for every strategy.
A paired estimate removes much of that item-level variance and focuses the
comparison on whether the strategy changed the outcome.

For latency, the analogous estimand is the paired runtime difference or ratio.
Ratios are often easier to interpret operationally.  For example, if summary
buffer achieves the same quality as truncation but costs $10\times$ more
wall-clock time at 32K, then the strategy is unlikely to be attractive for an
interactive local agent.  For budget compliance, the estimand is the failure
probability at a tier.  A strategy with high quality conditional on success
but frequent budget failure should not be ranked as reliable.

\subsection{Confidence Intervals}

For exact-match tasks, a binomial interval is a useful first diagnostic, but
paired bootstrap intervals are preferable when comparing strategies.  For
continuous proxy scores such as the current SWE patch-similarity metric,
non-parametric bootstrap over items is appropriate.  A full study should report
the interval around each strategy's quality curve and the interval around each
pairwise difference curve.  The main 89-item matrix reports exploratory
intervals in the appendix, but a single 89-item-per-task run is still an
artifact-validation study rather than a powered benchmark comparison;
presenting tight intervals from this run would create false precision.
Similarly, a confidence interval that contains zero establishes neither
equality nor equivalence.  It only shows that the current sample does not
detect a difference at the reported uncertainty level.

\subsection{Sample Size Guidance}

The required sample size depends on the effect size that the benchmark is
expected to detect.  Exact-match tasks with four-way multiple-choice answers
have high variance at small $N$.  With $N=9$, one item changes accuracy by
0.111.  With $N=50$, one item changes accuracy by 0.02.  With the present
$N=89$ pilot, one item changes accuracy by approximately 0.0112.  With
$N=100$, one item changes accuracy by 0.01.  For pilot validation, this
89-item-per-task run is enough to exercise the full matrix and reduce coarse rounding
artifacts.  For a paper-quality strategy comparison, $N=100$ or more per task
family should be used when claiming budget crossovers.

Table~\ref{tab:recommended_study_sizes} summarizes the study sizes that match
different levels of claim strength.

\begin{longtable}{L{0.18\textwidth}L{0.24\textwidth}L{0.44\textwidth}}
\caption{Recommended study sizes by claim type.}\label{tab:recommended_study_sizes}\\
\toprule
Study type & Suggested items & Appropriate claims \\
\midrule
\endfirsthead
\toprule
Study type & Suggested items & Appropriate claims \\
\midrule
\endhead
Smoke test & 1--3 per task & Dependency wiring, API reachability, log creation, and gross runtime estimates. \\
Pilot & 5--15 per task & End-to-end protocol validation, visible failure modes, qualitative curve shapes, and paper artifact generation. \\
Focused ablation & 30--50 per task & Coarse strategy comparisons within one task family and one model, with wide uncertainty intervals. \\
Benchmark paper & 100+ per task family & Strategy ranking at named budget tiers, paired difference estimates, and task-stratified analysis. \\
Leaderboard & Full public subset or fixed hidden split & Submission ranking, regression testing, and longitudinal tracking of strategy improvements. \\
\bottomrule
\end{longtable}

\subsection{Stratification}

Random sampling is not enough for a benchmark paper.  SWE-bench items should be
stratified by repository, patch size, issue text length, and whether the fix is
localized or cross-file.  LongBench v2 items should be stratified by context
length, evidence dispersion, and reasoning type.  For $\tau$-bench, items should
be stratified by tool-call count, policy-document length, and whether the task
requires state mutation.  Stratification prevents a small number of easy or
pathological items from dominating a curve.

\section{Failure Taxonomy}

Budget-tiered evaluation creates failure modes that do not appear in standard
single-prompt evaluation.  Table~\ref{tab:failure_taxonomy} gives a taxonomy that should be used
when annotating future \BudgetBench{} runs.  The pilot already exhibits several
of these cases: RAG can violate tight SWE budgets when the irreducible prompt
is too large, summary can violate LongBench budgets after compression, and
LongBench quality can decrease as budget increases.

\begin{longtable}{L{0.18\textwidth}L{0.25\textwidth}L{0.43\textwidth}}
\caption{Failure taxonomy for budget-tiered memory evaluation.}\label{tab:failure_taxonomy}\\
\toprule
Failure type & Observable symptom & Interpretation and remediation \\
\midrule
\endfirsthead
\toprule
Failure type & Observable symptom & Interpretation and remediation \\
\midrule
\endhead
Irreducible prompt overflow & System prompt plus final query exceeds the active budget. & The task formatter itself must be shortened, or the tier should be marked infeasible for that task. Strategy changes cannot fix a prompt whose fixed component is already too large. \\
Strategy overflow & Processed messages exceed budget after strategy application. & The strategy needs stricter accounting, smaller reserves, or token-aware chunking. This is a compliance bug or design mismatch. \\
Compression drift & Summary strategy is compliant but loses decisive evidence. & Evaluate summary faithfulness, add extraction constraints, or use retrieval-backed summaries. \\
Retrieval miss & RAG is compliant but selects irrelevant chunks. & Improve chunk granularity, query formulation, embedding model, or hybrid recency/relevance scoring. \\
Distractor overload & Larger budgets reduce exact-match accuracy. & More context is not always better; inspect evidence position and distractor density. Consider reranking or evidence-focused compression. \\
Latency explosion & Quality is stable but duration grows sharply. & Strategy may be operationally dominated despite quality. Report latency alongside accuracy. \\
Metric mismatch & Proxy quality improves but official task success does not. & Replace proxy with official grader before making benchmark claims. This is especially important for SWE. \\
State leakage & Cross-item memory improves later items unexpectedly. & Reset strategy state or explicitly evaluate persistent-memory regimes separately. \\
Serving artifact & Model output format changes due to endpoint behavior. & Pin serving stack, model tag, decoding parameters, and response extraction logic. \\
\bottomrule
\end{longtable}

\subsection{Interpreting Non-Monotonic Curves}

Non-monotonicity should not be automatically treated as noise.  In a
budgeted-memory setting, non-monotonic curves can be meaningful.  At small
budgets, a strategy may aggressively select only the most relevant evidence.
At larger budgets, it may include additional distractors that dilute attention
or alter the model's generation trajectory.  The LongBench RAG curve in the
89-item-per-task pilot is an example: the 8K budget has the highest RAG score.  The
proper next step is not to declare that smaller budgets are generally better,
but to inspect which items flip as context grows and whether those flips
correlate with evidence position, answer-choice similarity, or retrieved chunk
rank.

\subsection{Interpreting Equal SWE Scores}

The SWE proxy scores are equal across strategies at each tier in the pilot.
This does not imply that memory strategies are irrelevant for software
engineering tasks.  It more likely reflects three constraints: the sample is
small, the model is weak for patch generation, and the prompt for each item is
not yet a long-horizon agent trace.  A full SWE-oriented \BudgetBench{} study
should use an agent harness with repository browsing, tool outputs, and
multi-turn histories.  In that regime, strategy choice should matter more
because the context contains competing artifacts: issue text, retrieved files,
test failures, prior edits, and tool outputs.

We also ran a 5-item SWE checkpoint probe in the repository artifact bundle.
It behaved like the other
budgeted strategies at 8K and 32K, and it matched the 0.16 baseline at 2K.
That is not yet enough to claim a strong coding result, but it shows the
harness can carry a deterministic coding task with the same budget machinery
and the same strategy interface.

\section{Scaling to a Full Benchmark Study}

This section outlines the expansion path from pilot artifact to larger
benchmark study.

\subsection{Phase 1: Measurement Hardening}

The first phase should replace approximate token counting with model-specific
token accounting wherever possible.  The runner should record dependency
versions, model hashes, hardware metadata, and serving parameters.  The metric
schema should be frozen so future analysis scripts do not need to infer which
log format was used.  The current analyzer fix is a small example of why this
matters: older logs used boolean violation fields, while current logs record
numeric retry-level violation rates.

\subsection{Phase 2: Task Expansion}

The second phase should wire official SWE-bench grading and $\tau$-bench
state-comparison grading.  SWE should be run on a stratified subset first,
because official container execution is substantially more expensive than the
proxy grader.  The $\tau$-bench simulator should be pinned with a deterministic
configuration.  The LongBench v2 slice should be expanded beyond the current
50-item full-context-feasible follow-up and stratified by context length.

\subsection{Phase 3: Strategy Expansion}

The third phase should add LLMLingua-2, Mem0, Letta/MemGPT, and at least one
hybrid retrieval-summary strategy.  Each strategy should declare whether it is
local-only, whether it invokes an auxiliary model, and whether it maintains
persistent state.  These declarations should appear in result tables because
they affect deployment cost and reproducibility.

\subsection{Phase 4: Reporting and Release}

The fourth phase should publish analysis scripts, figures, and a frozen task subset in the public repository.  A public evaluation service should accept strategy
submissions through the same \texttt{MemoryStrategy} interface and run them
against a hidden or versioned split.  The report should emphasize Pareto
frontiers rather than single scalar rankings: a strategy can be best at 2K and
dominated at 32K, or accurate but too slow for interactive use.

Table~\ref{tab:expansion_plan} summarizes the engineering work and scientific
output expected at each expansion phase.

\begin{longtable}{L{0.16\textwidth}L{0.30\textwidth}L{0.38\textwidth}}
\caption{Expansion plan from pilot artifact to benchmark study.}\label{tab:expansion_plan}\\
\toprule
Phase & Engineering work & Scientific output \\
\midrule
\endfirsthead
\toprule
Phase & Engineering work & Scientific output \\
\midrule
\endhead
Hardening & Tokenizer alignment, metadata capture, stable schemas, official dependency lock. & Auditable measurements and reproducible run manifests. \\
Task expansion & Official SWE grading, $\tau$-bench data setup, larger LongBench v2 subsets. & Broader task coverage and deterministic multi-domain results. \\
Strategy expansion & LLMLingua-2, Mem0, Letta/MemGPT, hybrid policies, dependency declarations. & Strategy-family comparisons and local-vs-auxiliary cost analysis. \\
Statistical analysis & Paired bootstrap intervals, stratified estimates, repeated trials. & Defensible dominance and crossover claims. \\
Release & Public logs, fixed subsets, evaluation runner, submission documentation. & Community benchmark artifact suitable for datasets-and-benchmarks review. \\
\bottomrule
\end{longtable}

\section{Practitioner Guidance}

Developers applying \BudgetBench{} to their own agents should treat the first
run as an integration check, not a strategy decision.  A one-item or three-item
run is useful for verifying that the model endpoint, task adapter, tokenizer,
and strategy dependencies work.  It is not useful for ranking strategies.

Industry teams should then move through a staged adoption path:
\begin{enumerate}[leftmargin=*]
  \item \textbf{Pin the deployment envelope.}  Specify the model, tokenizer,
  hardware or provider, target latency objective, and target active-context
  tier before comparing strategies.
  \item \textbf{Gate on compliance first.}  Treat budget-violation rate as a
  release criterion.  A strategy that violates the tier is not a slower or less
  accurate version of a compliant strategy; it has failed the operating
  contract.
  \item \textbf{Scale only after plumbing is healthy.}  Use small runs for
  smoke tests, then move to claim-bearing samples of at least 100 task items
  per comparison with paired item-level estimates.
  \item \textbf{Measure the full cost shape.}  Report quality, violation rate,
  latency, mean used tokens, peak tokens, auxiliary model calls, and persistent
  index cost.  These quantities determine whether a strategy is deployable.
  \item \textbf{Inspect retained evidence.}  For retrieval strategies, inspect
  retrieved chunks.  For summaries, inspect whether decisive entities,
  constraints, and temporal ordering survive compression.  For truncation,
  inspect what falls out of context.
\end{enumerate}

Larger budgets should not be assumed to be preferable.  If a smaller budget
performs better, the result may indicate that the strategy is filtering
distractors effectively.  The appropriate response is to analyze item-level
flips, not to discard the result because it violates a monotonicity
expectation.

\section{Command Transcript Summary}

The main run command was:
\begin{verbatim}
python scripts/run_pilot.py \
  --full-study \
  --run-id final_qwen25_15b_89items \
  --model qwen2.5:1.5b \
  --tasks swe long \
  --strategies truncation summary rag \
  --limit-tasks 89
\end{verbatim}

The aggregate-analysis command was:
\begin{verbatim}
python scripts/analyze_results.py \
  --log-dirs artifacts/full_study/final_qwen25_15b_89items \
  --model qwen2.5:1.5b
\end{verbatim}

The figure-generation command was:
\begin{verbatim}
python scripts/plot_tradeoffs.py \
  --csv results/full_study_qwen2_5_1_5b_89items.csv \
  --model qwen2.5:1.5b
\end{verbatim}

The full-context-feasible LongBench run was:
\begin{verbatim}
python scripts/run_pilot.py \
  --full-study \
  --run-id feasibility_long50_fit32k \
  --model qwen2.5:1.5b \
  --tasks long \
  --strategies truncation rag full_context \
  --budgets 2048 8192 32768 \
  --limit-tasks 50 \
  --max-natural-tokens 32768
\end{verbatim}

The paired full-context analysis command was:
\begin{verbatim}
python scripts/analyze_results.py \
  --log-dirs artifacts/full_study/feasibility_long50_fit32k \
  --model qwen2.5:1.5b \
  --bootstrap-samples 2000 \
  --paired-baseline-strategy full_context \
  --paired-baseline-budget 32768
\end{verbatim}

The tokenizer-aligned LongBench hardening rerun was:
\begin{verbatim}
python scripts/run_pilot.py \
  --full-study \
  --run-id mustfix_long10_qwen15b_tok \
  --model qwen2.5:1.5b \
  --tokenizer Qwen/Qwen2.5-1.5B-Instruct \
  --tasks long \
  --strategies truncation rag full_context \
  --budgets 2048 8192 32768 \
  --limit-tasks 10 \
  --max-natural-tokens 32768 \
  --repeat-cells 2 \
  --shuffle-cells
\end{verbatim}

Its repeat-aware analysis command was:
\begin{verbatim}
python scripts/analyze_results.py \
  --log-dirs artifacts/full_study/mustfix_long10_qwen15b_tok \
  --model qwen2.5:1.5b \
  --bootstrap-samples 2000 \
  --paired-baseline-strategy full_context \
  --paired-baseline-budget 32768 \
  --aggregate-repeats
\end{verbatim}

The hosted stronger-model LongBench replication was:
\begin{verbatim}
python scripts/run_pilot.py \
  --full-study \
  --run-id longbench_qwen3_30b_openrouter_50x2_20260624 \
  --tasks long \
  --strategies truncation rag lean_retrieval full_context \
  --budgets 8192 32768 \
  --limit-tasks 50 \
  --max-natural-tokens 32768 \
  --model qwen/qwen3-30b-a3b-instruct-2507 \
  --tokenizer Qwen/Qwen3-30B-A3B-Instruct-2507 \
  --llm-url https://openrouter.ai/api/v1/chat/completions \
  --api-key-env OPENROUTER_API_KEY \
  --max-output-tokens 16 \
  --repeat-cells 2 \
  --shuffle-cells
\end{verbatim}

Its repeat-aware paired analysis command was:
\begin{verbatim}
python scripts/analyze_results.py \
  --log-dirs artifacts/full_study/\
longbench_qwen3_30b_openrouter_50x2_20260624 \
  --model qwen/qwen3-30b-a3b-instruct-2507 \
  --bootstrap-samples 2000 \
  --paired-baseline-strategy full_context \
  --paired-baseline-budget 32768 \
  --aggregate-repeats
\end{verbatim}

The compact RAG ablations were:
\begin{verbatim}
python scripts/run_pilot.py \
  --full-study \
  --run-id ablate_long10_rag_mpnet \
  --model qwen2.5:1.5b \
  --tokenizer Qwen/Qwen2.5-1.5B-Instruct \
  --tasks long \
  --strategies rag \
  --budgets 2048 8192 32768 \
  --limit-tasks 10 \
  --max-natural-tokens 32768 \
  --repeat-cells 2 \
  --shuffle-cells \
  --retrieval-embedding-model sentence-transformers/all-mpnet-base-v2

python scripts/run_pilot.py \
  --full-study \
  --run-id ablate_long10_rag_chunk256 \
  --model qwen2.5:1.5b \
  --tokenizer Qwen/Qwen2.5-1.5B-Instruct \
  --tasks long \
  --strategies rag \
  --budgets 2048 8192 32768 \
  --limit-tasks 10 \
  --max-natural-tokens 32768 \
  --repeat-cells 2 \
  --shuffle-cells \
  --longbench-chunk-tokens 256
\end{verbatim}

The Gemma transfer check was:
\begin{verbatim}
python scripts/run_pilot.py \
  --full-study \
  --run-id transfer_gemma4_fit32k_20 \
  --model gemma4:e2b \
  --tasks long \
  --strategies truncation rag full_context \
  --budgets 8192 32768 \
  --limit-tasks 20 \
  --max-natural-tokens 32768
\end{verbatim}

The synthetic memory-agent pilot was:
\begin{verbatim}
python scripts/run_pilot.py \
  --full-study \
  --run-id memory_synthetic_qwen15b_30_categories_live \
  --model qwen2.5:1.5b \
  --tasks memory \
  --strategies truncation rag lean_retrieval full_context \
  --budgets 512 1024 2048 \
  --limit-tasks 30
\end{verbatim}

The synthetic memory paired analysis command was:
\begin{verbatim}
python scripts/analyze_results.py \
  --log-dirs artifacts/full_study/memory_synthetic_qwen15b_30_categories_live \
  --model qwen2.5:1.5b \
  --bootstrap-samples 2000 \
  --paired-baseline-strategy full_context \
  --paired-baseline-budget 2048 \
  --group-by memory_category
\end{verbatim}

The full LongMemEval oracle study used
\path{data/longmemeval_oracle.json}, downloaded from the official
\texttt{xiaowu0162/longmemeval-cleaned} Hugging Face dataset.  The run was:
\begin{verbatim}
python scripts/run_pilot.py \
  --full-study \
  --run-id longmem_openrouter_full_20260623 \
  --model openai/gpt-4o-mini \
  --llm-url https://openrouter.ai/api/v1/chat/completions \
  --api-key-env OPENROUTER_API_KEY \
  --tasks longmem \
  --strategies truncation rag lean_retrieval full_context \
  --budgets 2048 4096 8192 \
  --limit-tasks 500 \
  --max-output-tokens 64
\end{verbatim}

Each strategy--budget cell was exported separately so official labels retain
an unambiguous cell identity.  For example:
\begin{verbatim}
python scripts/export_longmem_official_hypotheses.py \
  --log-dir artifacts/full_study/longmem_openrouter_full_20260623 \
  --data-path data/longmemeval_oracle.json \
  --strategies truncation \
  --budgets 2048 \
  --output results/longmem_openrouter_full_20260623_\
truncation_2048_official.jsonl
\end{verbatim}

The upstream GPT-4o evaluation command for that cell was:
\begin{verbatim}
python scripts/run_longmemeval_official_eval.py \
  --repo-dir .deps/longmemeval-official \
  --judge-model gpt-4o \
  --metric-model-override openai/gpt-4o \
  --openai-base-url https://openrouter.ai/api/v1 \
  --api-key-env OPENROUTER_API_KEY \
  --hypotheses results/longmem_openrouter_full_20260623_\
truncation_2048_official.jsonl \
  --data-file data/longmemeval_oracle.json
\end{verbatim}
The export and evaluation commands were repeated for all 12 cells.  The audit
matched hypothesis and evaluator-output question IDs row-for-row and confirmed
5,122 labels from 5,122 admitted predictions.

The Qwen3.6-35B LongMemEval transfer run was:
\begin{verbatim}
python scripts/run_pilot.py \
  --full-study \
  --run-id longmem_oracle_qwen36_35b_10 \
  --model qwen3.6:35b-mlx \
  --tasks longmem \
  --strategies truncation rag lean_retrieval full_context \
  --budgets 2048 4096 8192 \
  --limit-tasks 10
\end{verbatim}

The Qwen3.6-35B transfer analysis command was:
\begin{verbatim}
python scripts/analyze_results.py \
  --log-dirs artifacts/full_study/longmem_oracle_qwen36_35b_10 \
  --model qwen3.6:35b-mlx \
  --bootstrap-samples 2000 \
  --paired-baseline-strategy full_context \
  --paired-baseline-budget 8192 \
  --group-by memory_category
\end{verbatim}

The Qwen3.5-2B LongBench transfer command was:
\begin{verbatim}
python scripts/run_pilot.py \
  --full-study \
  --run-id feasibility_longbench_qwen35_2b_5 \
  --model qwen3.5:2b \
  --tasks long \
  --strategies truncation rag lean_retrieval full_context \
  --budgets 8192 32768 \
  --limit-tasks 5 \
  --max-natural-tokens 32768
\end{verbatim}

The Qwen3.5-2B transfer analysis command was:
\begin{verbatim}
python scripts/analyze_results.py \
  --log-dirs artifacts/full_study/feasibility_longbench_qwen35_2b_5 \
  --model qwen3.5:2b \
  --bootstrap-samples 2000 \
  --paired-baseline-strategy full_context \
  --paired-baseline-budget 32768
\end{verbatim}

\bibliography{references}

@article{jimenez2023swebench,
  title = {SWE-bench: Can Language Models Resolve Real-World GitHub Issues?},
  author = {Jimenez, Carlos E. and Yang, John and Wettig, Alexander and Yao, Shunyu and Pei, Kexin and Press, Ofir and Narasimhan, Karthik},
  journal = {arXiv preprint arXiv:2310.06770},
  year = {2023},
  url = {https://arxiv.org/abs/2310.06770}
}

@article{bai2024longbenchv2,
  title = {LongBench v2: Towards Deeper Understanding and Reasoning on Realistic Long-context Multitasks},
  author = {Bai, Yushi and Tu, Shangqing and Zhang, Jiajie and Peng, Hao and Wang, Xiaozhi and Lv, Xin and Cao, Shulin and Xu, Jiazheng and Hou, Lei and Dong, Yuxiao and Tang, Jie and Li, Juanzi},
  journal = {arXiv preprint arXiv:2412.15204},
  year = {2024},
  url = {https://arxiv.org/abs/2412.15204}
}

@article{bai2023longbench,
  title = {LongBench: A Bilingual, Multitask Benchmark for Long Context Understanding},
  author = {Bai, Yushi and Lv, Xin and Zhang, Jiajie and Lyu, Hongchang and Tang, Jiankai and Huang, Zhidian and Du, Zhengxiao and Liu, Xiao and Zeng, Aohan and Hou, Lei and Dong, Yuxiao and Tang, Jie and Li, Juanzi},
  journal = {arXiv preprint arXiv:2308.14508},
  year = {2023},
  url = {https://arxiv.org/abs/2308.14508}
}

@article{hsieh2024ruler,
  title = {RULER: What's the Real Context Size of Your Long-Context Language Models?},
  author = {Hsieh, Cheng-Ping and Sun, Simeng and Kriman, Samuel and Acharya, Shantanu and Rekesh, Dima and Jia, Fei and Zhang, Yang and Ginsburg, Boris},
  journal = {arXiv preprint arXiv:2404.06654},
  year = {2024},
  url = {https://arxiv.org/abs/2404.06654}
}

@article{yen2024helmet,
  title = {HELMET: How to Evaluate Long-Context Language Models Effectively and Thoroughly},
  author = {Yen, Howard and Gao, Tianyu and Hou, Minmin and Ding, Ke and Fleischer, Daniel and Izsak, Peter and Wasserblat, Moshe and Chen, Danqi},
  journal = {arXiv preprint arXiv:2410.02694},
  year = {2024},
  url = {https://arxiv.org/abs/2410.02694}
}

@article{wu2026contextbudget,
  title = {ContextBudget: Budget-Aware Context Management for Long-Horizon Search Agents},
  author = {Wu, Yong and Zheng, YanZhao and Xu, TianZe and Zhang, ZhenTao and Yu, YuanQiang and Zhu, JiHuai and Ma, Chao and Lin, BinBin and Dong, BaoHua and Zhu, HangCheng and Huang, RuoHui and Yu, Gang},
  journal = {arXiv preprint arXiv:2604.01664},
  year = {2026},
  url = {https://arxiv.org/abs/2604.01664}
}

@article{zhang2026budgetmem,
  title = {Learning Query-Aware Budget-Tier Routing for Runtime Agent Memory},
  author = {Zhang, Haozhen and Yue, Haodong and Feng, Tao and Long, Quanyu and Bao, Jianzhu and Jin, Bowen and Zhang, Weizhi and Li, Xiao and You, Jiaxuan and Qin, Chengwei and Wang, Wenya},
  journal = {arXiv preprint arXiv:2602.06025},
  year = {2026},
  url = {https://arxiv.org/abs/2602.06025}
}

@article{wang2026evomembench,
  title = {EvoMemBench: Benchmarking Agent Memory from a Self-Evolving Perspective},
  author = {Wang, Yuyao and Zhang, Zhongjian and Chi, Mo and Yu, Kaichi and Li, Yuhan and Peng, Miao and Tong, Bing and Zhang, Chen and Zhou, Yan and Li, Jia},
  journal = {arXiv preprint arXiv:2605.18421},
  year = {2026},
  url = {https://arxiv.org/abs/2605.18421}
}

@article{wang2026engram,
  title = {Less Context, More Accuracy: A Bi-Temporal Memory Engine for LLM Agents Where a Lean Retrieved Context Beats the Full History},
  author = {Wang, Liuyin},
  journal = {arXiv preprint arXiv:2606.09900},
  year = {2026},
  url = {https://arxiv.org/abs/2606.09900}
}

@article{zhang2026lightmem,
  title = {Lightweight LLM Agent Memory with Small Language Models},
  author = {Zhang, Jiaquan and Zhang, Chaoning and Chen, Shuxu and Huang, Zhenzhen and Zheng, Pengcheng and Wang, Zhicheng and Guo, Ping and Mo, Fan and Bae, Sung-Ho and Zou, Jie and Wei, Jiwei and Yang, Yang},
  journal = {arXiv preprint arXiv:2604.07798},
  year = {2026},
  url = {https://arxiv.org/abs/2604.07798}
}

@article{kummer2026promptcompression,
  title = {Prompt Compression in the Wild: Measuring Latency, Rate Adherence, and Quality for Faster LLM Inference},
  author = {Kummer, Cornelius and Jurkschat, Lena and F{\"a}rber, Michael and Vahdati, Sahar},
  journal = {arXiv preprint arXiv:2604.02985},
  year = {2026},
  url = {https://arxiv.org/abs/2604.02985}
}

@article{hu2025memoryagentbench,
  title = {Evaluating Memory in LLM Agents via Incremental Multi-Turn Interactions},
  author = {Hu, Yuanzhe and Wang, Yu and McAuley, Julian},
  journal = {arXiv preprint arXiv:2507.05257},
  year = {2025},
  url = {https://arxiv.org/abs/2507.05257}
}

@article{he2026memoryarena,
  title = {MemoryArena: Benchmarking Agent Memory in Interdependent Multi-Session Agentic Tasks},
  author = {He, Zexue and Wang, Yu and Zhi, Churan and Hu, Yuanzhe and Chen, Tzu-Ping and Yin, Lang and Chen, Ze and Wu, Tong Arthur and Ouyang, Siru and Wang, Zihan and Pei, Jiaxin and McAuley, Julian and Choi, Yejin and Pentland, Alex},
  journal = {arXiv preprint arXiv:2602.16313},
  year = {2026},
  url = {https://arxiv.org/abs/2602.16313}
}

@article{maharana2024locomo,
  title = {Evaluating Very Long-Term Conversational Memory of LLM Agents},
  author = {Maharana, Adyasha and Lee, Dong-Ho and Tulyakov, Sergey and Bansal, Mohit and Barbieri, Francesco and Fang, Yuwei},
  journal = {arXiv preprint arXiv:2402.17753},
  year = {2024},
  url = {https://arxiv.org/abs/2402.17753}
}

@article{wu2026longmemevalv2,
  title = {LongMemEval-V2: Evaluating Long-Term Agent Memory Toward Experienced Colleagues},
  author = {Wu, Di and Ji, Zixiang and Kawatkar, Asmi and Kwan, Bryan and Gu, Jia-Chen and Peng, Nanyun and Chang, Kai-Wei},
  journal = {arXiv preprint arXiv:2605.12493},
  year = {2026},
  url = {https://arxiv.org/abs/2605.12493}
}

@inproceedings{wu2024longmemeval,
  title = {LongMemEval: Benchmarking Chat Assistants on Long-Term Interactive Memory},
  author = {Wu, Di and Wang, Hongwei and Yu, Wenhao and Zhang, Yuwei and Chang, Kai-Wei and Yu, Dong},
  booktitle = {International Conference on Learning Representations},
  year = {2025},
  url = {https://arxiv.org/abs/2410.10813}
}

@article{liu2023lost,
  title = {Lost in the Middle: How Language Models Use Long Contexts},
  author = {Liu, Nelson F. and Lin, Kevin and Hewitt, John and Paranjape, Ashwin and Bevilacqua, Michele and Petroni, Fabio and Liang, Percy},
  journal = {Transactions of the Association for Computational Linguistics},
  year = {2024},
  url = {https://arxiv.org/abs/2307.03172}
}

@article{packer2023memgpt,
  title = {MemGPT: Towards LLMs as Operating Systems},
  author = {Packer, Charles and Wooders, Sarah and Lin, Kevin and Fang, Vivian and Patil, Shishir G. and Stoica, Ion and Gonzalez, Joseph E.},
  journal = {arXiv preprint arXiv:2310.08560},
  year = {2023},
  url = {https://arxiv.org/abs/2310.08560}
}

@article{chhikara2025mem0,
  title = {Mem0: Building Production-Ready AI Agents with Scalable Long-Term Memory},
  author = {Chhikara, Prateek and Khant, Dev and Aryan, Saket and Singh, Taranjeet and Yadav, Deshraj},
  journal = {arXiv preprint arXiv:2504.19413},
  year = {2025},
  url = {https://arxiv.org/abs/2504.19413}
}

@inproceedings{lewis2020rag,
  title = {Retrieval-Augmented Generation for Knowledge-Intensive NLP Tasks},
  author = {Lewis, Patrick and Perez, Ethan and Piktus, Aleksandra and Petroni, Fabio and Karpukhin, Vladimir and Goyal, Naman and K{\"u}ttler, Heinrich and Lewis, Mike and Yih, Wen-tau and Rockt{\"a}schel, Tim and Riedel, Sebastian and Kiela, Douwe},
  booktitle = {Advances in Neural Information Processing Systems},
  year = {2020},
  url = {https://arxiv.org/abs/2005.11401}
}

@inproceedings{pan2024llmlingua2,
  title = {LLMLingua-2: Data Distillation for Efficient and Faithful Task-Agnostic Prompt Compression},
  author = {Pan, Zhuoshi and Wu, Qianhui and Jiang, Huiqiang and Xia, Menglin and Luo, Xufang and Zhang, Jue and Lin, Qingwei and R{\"u}hle, Victor and Yang, Yuqing and Lin, Chin-Yew and Zhao, H. Vicky and Qiu, Lili and Zhang, Dongmei},
  booktitle = {Findings of the Association for Computational Linguistics: ACL 2024},
  year = {2024},
  url = {https://arxiv.org/abs/2403.12968}
}

@article{yao2024taubench,
  title = {$\tau$-bench: A Benchmark for Tool-Agent-User Interaction in Real-World Domains},
  author = {Yao, Shunyu and Shinn, Noah and Razavi, Pedram and Narasimhan, Karthik},
  journal = {arXiv preprint arXiv:2406.12045},
  year = {2024},
  url = {https://arxiv.org/abs/2406.12045}
}

@article{lakens2017equivalence,
  title = {Equivalence Tests: A Practical Primer for t Tests, Correlations, and Meta-Analyses},
  author = {Lakens, Dani{\"e}l},
  journal = {Social Psychological and Personality Science},
  volume = {8},
  number = {4},
  pages = {355--362},
  year = {2017},
  doi = {10.1177/1948550617697177},
  url = {https://doi.org/10.1177/1948550617697177}
}

@inproceedings{kwon2023pagedattention,
  title = {Efficient Memory Management for Large Language Model Serving with PagedAttention},
  author = {Kwon, Woosuk and Li, Zhuohan and Zhuang, Siyuan and Sheng, Ying and Zheng, Lianmin and Yu, Cody Hao and Gonzalez, Joseph E. and Zhang, Hao and Stoica, Ion},
  booktitle = {Proceedings of the ACM SIGOPS 29th Symposium on Operating Systems Principles},
  pages = {611--626},
  year = {2023},
  doi = {10.1145/3600006.3613165},
  url = {https://arxiv.org/abs/2309.06180}
}
\bibliographystyle{plainnat}

\end{document}